\newif\ifprint

\newif\ifdraft
\newif\ifArxiv
\newif\ifJDEDS

\printtrue

\Arxivtrue
\JDEDStrue

\PassOptionsToClass{class/jdeds}{my-class/my-class}

\documentclass{my-class/my-class}

\usepackage{my-goodcharter}
\usepackage{microtype}

\usepackage{mathtools,amssymb}

\usepackage{bm}

\usepackage{my-semantic-logic-symbols}
\usepackage{my-bnf}
\usepackage{my-linewise}

\usepackage{delarray}

\usepackage[
  shortcuts,
  allowbreakbefore,
]{extdash}

\usepackage[
  sort,
]{natbib}

\usepackage[inline]{enumitem}
\usepackage{tasks}

\usepackage{bigdelim}

\usepackage{subcaption}

\usepackage{hyperref}
\usepackage{stringenc}
\makeatletter
\AtEndPreamble{\hypersetup{ % AfterPreamble is too late for ArXiv for some reason
  pdfauthor={K. Ritsuka},
  pdfencoding=auto,
  psdextra
}}
\makeatother

\usepackage[open,openlevel=2]{bookmark}

\usepackage{orcidlink}
\usepackage{uri}

\ifdraft
  \renewcommand*{\backref}[1]{}
  \renewcommand*{\backrefalt}[4]{{\tiny[%
      \ifcase #1 Not cited.%
            \or Page~#2.%
            \else Pages #2.%
      \fi%
      ]}}

  \newcommand{\myBibPreambleText}{[Citing pages are listed after each reference.]}

  \makeatletter
  \renewcommand{\bibsection}{
    \section*{\bibname \\ {\normalsize\mdseries \myBibPreambleText}} % <------ choose `chapter' vs. `section' accordingly
    \@mkboth {\MakeUppercase {\bibname }}{\MakeUppercase {\bibname }}
  }
  \makeatother
\fi

\ifprint
  \PassOptionsToPackage{mycolor=print}{my-cactus}
\else
  \PassOptionsToPackage{mycolor=dark}{my-cactus}
\fi
\usepackage[
  mycoloroverride
]{my-cactus}

\usepackage{my-thm}

\usepackage{graphicx}
\usepackage{xcolor}

\mytitle{
  Do What You Know: Coupling Knowledge with Action in Discrete-Event Systems
}
\mydate{\today}
\mydedicatory{
  This work was financially supported
  by the Natural Sciences and Engineering Research Council of Canada (NSERC)
  through a Collaborative Research and Development Grant
  together with General Dynamics Land Systems\--Canada
  and Defence Research and Development Canada (K. Ritsuka)
  and an NSERC Discovery Grant (K. Rudie).
}

\myauthor{K. Ritsuka}
\myorcid{0000-0003-3713-5964}
\myemail{Ritsuka314@queensu.ca}
\myaddressref{qu}

\myauthor{Karen Rudie}
\myorcid{0000-0002-8675-334X}
\myemail{karen.rudie@queensu.ca}
\myaddressref{qu}

\myaddress{qu}{
  Department of Electrical and Computer Engineering and Ingenuity Labs Research Institute,\\
  Queen's University, Kingston, ON, Canada K7L 3N6
}

\mykeywords{Discrete\-/Event Systems \and Supervisory Control \and Epistemic Logic}
\mysubjclass{93A14 \and 93A16 \and 93B07 \and 93C65}

\myabstract{
  An epistemic model for decentralized discrete-event systems with non-binary control is presented.
  This framework combines existing work on conditional control decisions with existing work on formal reasoning about knowledge in discrete-event systems.
  The novelty in the model presented is that the necessary and sufficient conditions for problem solvability encapsulate the actions that supervisors must take.
  This direct coupling between knowledge and action
  \--- in a formalism that mimics natural language \---
  makes it easier,
  when the problem conditions fail,
  to determine how the problem requirements should be revised.
}

\begin{document}
\bstctlcite{BSTcontrol}

\mymaketitle

\section{Introduction}

The recent emergence of the Internet of things,
including smart vehicles,
home automation,
and wearables
has raised the need for decentralized supervisory control:
the concept that the control is performed by not a monolithic,
but many individual entities \--- or agents \--- separated by the environment.
This paper focuses on systems modelled as discrete-event systems (DES).

With control actions performed jointly,
a mechanism \--- called a \emph{fusion rule} \--- is needed to combine control decisions of the agents.
Decentralized control of discrete-event systems under partial observations
began with allowing only Boolean control decisions,
and synthesis of the control policy has been studied when the fusion rule is conjunctive \citep{Cieslak1988, Rudie1992},
and later generalized to other fusion rules \citep{Yoo2002}.
Further work by \citet{Yoo2004} extended the approach to allow non-binary control decisions with a more sophisticated fusion rule,
so that supervisors can ``conditionally'' turn on/off events based on the actions of other supervisors.
\citeauthor{Yoo2004} gave necessary and sufficient conditions for the existence of supervisors \citep{Yoo2004}
and a realization of the supervisors \citep{Yoo2005}.
% The realization requires the construction of four automata,
% each dedicated to one kind of control decision.
% Hence their realization can be seen as four Moore machines running in parallel.
% Their process of verifying whether supervisors exist,
% that of partitioning controllable events,
% and that of constructing the supervisors,
% are separate,
% as well as proofs of their respective correctness.

With a different approach,
\citet{Ricker2007} gave an epistemic interpretation of the work of \citet{Yoo2004},
where the use of the formal language of epistemic logic enabled one to
discuss the supervisory control in an anthropomorphic manner,
which gives a more intuitive understanding for how control decisions are made.
The epistemic logic model developed by \citet{Ricker2007}
resolves the first drawback of the linguistic approach listed above,
namely the meaning of an epistemic expression is immediately understandable at a glance,
so that an expression of the form $K_1(\phi)$ means ``Supervisor 1 knows $\phi$''.
However, even in the earlier epistemic logic reformulation of DES problems,
there is a tenuous connection between the solvability conditions
and the actions to be prescribed for supervisors
in a construction that exploits the conditions.

This paper extends our earlier work \citep{Ean2021a},
in which we gave epistemic logical characterizations
to architectures by \citet{Cieslak1988}, \citet{Rudie1992}, and \citet{Yoo2002}.
Following the same methodology, we
demonstrate and prove that a standard and representative result \citep{Yoo2004}
in decentralized DES can not only be cast as epistemic logic
but also in a way that results in a direct link
between the condition that must hold for a solution to exist
and the control protocol that must be followed when the condition holds.
In particular, we provide a line-by-line correspondence
between the expressions of the knowledge the supervisors must possess
and the actions they must take.

We also point out that epistemic expressions
could facilitate simpler formal discussions.
The verification of supervisor existence,
and the algorithm of supervisor synthesis,
are described by a single proof
which also demonstrate their correctness.
The two processes,
together with that of partitioning controllable events,
run as a coroutine.

We provide more comprehensive intuition
for the necessary and sufficient conditions
for the solvability and solution to a decentralized control problem,
including an analysis of why supervisors issuing \textbf{don't know} decisions
nonetheless might still ``know'' something,
and thus again demonstrate how the use of epistemic logic
provides a better understanding of the decentralized DES control problem.

Our treatment of the problem involves demonstrating
that if the control problems discussed here are solvable by supervisors of any flavour,
the problems must also be solvable by our knowledge-based supervisors.
This fact suggests that although the supervisors we construct are algorithmic,
since the algorithm is described in epistemic terms,
they reflect human-like reasoning.

\section{Preliminaries}

To lay a foundation for the discussion,
we recall the definition of discrete\-/event systems,
in particular that of the Decentralized Supervisory Control and Observation Problem,
then that of epistemic logic and its use in expressing inference\-/observability.
In this section we recall
the methodology we used in our previous work \citep{Ean2021a}.

\subsection{Discrete Event Systems}

The systems under consideration in this work are \textit{discrete-event systems}.
A discrete\-/event system is
a system with a discrete state space, where actions or event occurrences may cause the system to change state. We consider the system's behaviours to be all finite sequences of events the system can generate from its initial state.

In this work, we use the formalisms of discrete\-/event systems found in
\citet{Wonham2018} and
\citet{Cassandras2007}.

\begin{defn}
  We model a \emph{plant} $G$ as a finite state automaton (FSA)
  \[
    G = (\Sigma, Q, \delta, q_0)
  \]
  where $\Sigma$ is a finite set of \emph{events},
  $Q$ a finite set of \emph{states},
  $\delta$ the \emph{transition function},
  and $q_0 \in Q$ the \emph{initial state}.

  When more than one automaton is under discussion, we put the name of the automaton as a superscript in the components, e.g., we use $Q^G$ to refer to the state set of $G$.

  The \emph{language generated by $G$} is defined as
  \[
    L(G) = \setcomp{s \in \Sigma^*}{\delta(q_0, s)!}
  \]
 \end{defn}
We interpret $L(G)$ as the set of physically possible behaviours of $G$.

  A language $L$ is \emph{prefix\-/closed} whenever for all strings $s\sigma \in L$,
  it is always the case that $s \in L$. By definition, $L(G)$ is always prefix\-/closed.

\subsection{Decentralized Supervisory Control with Partial Observations}

When the behaviour of a discrete-event system plant's is not desirable, we constrain its behaviours through supervisory control.  We will allow an arbitrary number of supervisors to jointly perform the control, where each supervisor observes and controls a subset of events, and may issue a control decision for a given event. Decentralized control has been examined by many DES researchers. For a more extensive discussion,
see the decentralized control section in \citet{Cassandras2007}.

Because multiple supervisors may act on any given event, we require a mechanism to combine the (potentially conflicting or differing) control decisions of the different supervisors. 
\citet{Prosser1997} named such mechanisms \emph{fusion rules}. Subsequent work by \citet{Yoo2002} recognized that fusion rules for each event can be chosen separately and independently.

% There would usually be some conditions to prevent supervisors from cheating.
% E.g., completeness requires supervisors not to turn off uncontrollable events,
% feasibility and validity require a supervisor to behave consistently upon
% the generating of two distinct sequences but perceived identically under the supervisor's partial observation.
% We choose to encode these requirements in the architecture
% and not state them separately,
% to simplify the discussions and proofs.

Formally, we define the decentralized supervisory architecture as follows.

\begin{defn}
  \label{defn:arch}

  Let $\CD$ be a set of supervisory control decisions.
  Let $\nset = \{f_1, \dots, f_n\}$ be a finite set of $n$ supervisors for plant $G$.
  For simplicity, we will write $i$ instead of $f_i$ when referring to the supervisor.

  For each supervisor $i \in \nset$,
  let $\Sigma_{i,c}, \Sigma_{i,o} \subseteq \Sigma$
  be the sets of controllable and observable events for supervisor $i$, resp.
  Let $P_i : \Sigma^* \to \Sigma_{i,o}^*$ be the usual projection function
  to capture a supervisor $i$'s observation,
  i.e.,
  if a plant generates a sequence of events $s$, supervisor $i$ will only see $P_i(s)$.
  Denote the set of events controlled by some supervisors
  $\Sigma_c = \Union_{i \in \nset} \Sigma_{i,c}$,
  and the set of events not controlled by any supervisor
  $\Sigma_{uc} = \Sigma - \Sigma_c$.
  $\Sigma_{uc} = \Sect_{i \in \nset} \Sigma - \Sigma_{i,c}$.
  Hence we have $\Sigma_{uc} = \Sigma - \Sigma_c$.
  The sets $\Sigma_o$ and $\Sigma_{uo}$ are defined similarly.  
  Let $\nset_\sigma = \setcomp{i \in \nset}{\sigma \in \Sigma_{i,c}}$
  be the set of supervisors that can control $\sigma$.

  With a slight abuse of notation, we use $P_i(G)$ to denote the automaton
  constructed by replacing all transitions labelled by an unobservable event with $\emptyseq$
  and determinized,
  so that $P_i(G)$ recognizes the language $P_iL(G)$.

  Now supervisors can be prescribed by
  $f_i : P_iL(G) \times \Sigma_{i,c} \to \CD$ for all $f_i \in \nset$.
  Specifying that supervisors take arguments from $P_iL(G)$ instead of $L(G)$ implicitly encodes requirements traditionally referred to as \emph{feasibility} and \emph{validity},
  i.e., a supervisor must make consistent decisions for strings $s, s'$ 
  that look alike to that supervisor,
  i.e., such that $P_i(s) = P_i(s')$.
  We focus only on FSA\-/based supervisors.
  That is,
  a supervisor $f_i$ can be realized as a Moore machine $(S_i, f'_i)$
  such that $f_i(s, \sigma) = f'_i(\delta_i(s, q_{i, 0}))$,
  where $S_i$ is an FSA $(\Sigma, Q_i, \delta_i, q_{i, 0})$,
  and $f'_i : Q_i \times \Sigma_{i,c} \to \CD$.
  We will refer to $f'_i$ simply as $f_i$ when convenient.

  For each controllable event $\sigma$,
  let $cd_{\nset_\sigma}$ denote the collection of control decisions issued by supervisors $i \in \nset_\sigma$,
  hence $cd_{\nset_\sigma}$ has exactly $|\nset_\sigma|$ elements.
  Let $\CD_{\nset_\sigma}$ be the collection of all such $cd_{\nset_\sigma}$'s.
  Let $\FD = \set{\fde ,\fdd}$ be the set of fused decisions.
  Let $f_\sigma : \CD_{\nset_\sigma} \to \FD$ be the fusion functions
  chosen separately for each $\sigma \in \Sigma_c$,
  and the joint supervision $f_\nset : L(G) \times \Sigma_c \to \FD$
  be defined as
  $f_\nset(s, \sigma) = f_\sigma(\setidx{f_i(P_i(s), \sigma)}{i \in \nset_\sigma})$.
  Consequently, only decisions issued by supervisors $i \in \nset_\sigma$ are fused,
  and decisions of supervisors not controlling event $\sigma$ are ignored.
  
  The closed\-/loop behaviour of the plant under joint supervision is
  denoted by $L(f_\nset / G)$,
  and defined inductively as the smallest set such that:
  
  \begin{itemize}
    \item $\emptyseq \in L(f_\nset / G)$
    \item $s \in L(f_\nset / G) \land s\sigma \in L(G) \land \sigma \in \Sigma_{uc} \implies s\sigma \in L(f_\nset / G)$
    \item $s \in L(f_\nset / G) \land s\sigma \in L(G) \land \sigma \in \Sigma_c \land f_\nset(s, \sigma) = \fde \implies s\sigma \in L(f_\nset / G)$
  \end{itemize}

  The second bullet point in the definition of closed-loop behaviour encodes the requirement traditionally referred to as \emph{completeness}:
  a physically possible event that is not controllable by any supervisor must be allowed to occur under supervision. The third bullet point says that a physically possible event that is controllable and for which the fused decision is \textbf{enable} must be allowed to occur under supervision.

  % \cref{fig:arch} illustrates the architecture defined above,
  % which is adapted from Fig. 2 by \citet{Yoo2004} with entities labelled according to our symbolism.

  % \begin{figure}[htbp]
  %   \centerline{
  %     \color{black}
  %     \includegraphics[width=\columnwidth]{./arch/arch.pdf}
  %   }
  %   \caption{Architecture of Decentralized Control of DES with Non-binary Control Decisions.}
  %   \label{fig:arch}
  % \end{figure}

\end{defn}

Whereas the fusion function $f$ can be seen as an n\-/ary operation on supervisory control decisions $\CD$,
there is no operation over the fused decision set $\FD$,
since elements in this set are to be interpreted as fused decisions and should be regarded as final. In particular,
whereas we may take $\CD = \FD$ as Boolean values and $f$ as a Boolean function
as existing works commonly do when it is convenient \citep{Rudie1992, Yoo2002},
when moving to non\-/binary control decisions
\citep{Yoo2005},
we clearly separate the two sets and hence $\FD$ should not be considered as Boolean values
(although still binary).
For this reason, we also do not use the two symbols $0, 1$ for elements of either set.

The sets $\CD$ and $\FD$ being disjoint
also simplifies discussion:
we can now refer to an element of either set without explicitly stating from which set it comes.
We also refer to a particular element of either set simply as a decision when no confusion would arise.

\begin{rmk}
  \label{rmk:cd-f}
  Whereas the set $\CD$ determines the number of distinct control decisions available to the supervisors,
  what those decisions mean
  \--- their semantics \---
  is given by the fusion rule $f$.  Nonetheless, although the symbols we choose for  control decisions may be formally meaningless, we will still choose them with the intended fusion rule in mind. For example, in what follows, we will use the symbol \textbf{on} (resp., \textbf{off}) as an element of $\CD$ with the intended meaning that some supervisor's decision is that an event should be allowed to occur (resp., not allowed to occur).
\end{rmk}

Constructing multiple supervisors jointly restricting a plant's behaviours
will be called the \emph{Decentralized Supervisory Control and Observation Problem} (DSCOP).
We will use the term ``condition'' (without qualification) to refer to the \emph{necessary and sufficient} condition needed to solve DSCOP.

For the sake of comparison,
we will use the following generic definition of DSCOP
as a common ground for subsequent discussions.

\begin{prob}[Decentralized Supervisory Control and Observation Problem, DSCOP]
  Given a plant $G$,
  % a prefix\-/closed sublanguage of $L(G)$ arranged without loss of generality to be recognized by 
  a subautomaton\footnote{
    Our formulation is not restricted by requiring
    that $E$ is a subautomaton of $G$,
    since given an arbitrary $G'$ and $E'$ where $L(E') \subseteq L(G')$,
    one can always find language-equivalent $G$ and $E$
    such that $L(G) = L(G')$ and $L(E) = L(E')$
    and $E$ is a subautomaton of $G$.
  } $E$ of $G$, and
  $n$ pairs of controllable/observable event sets,
  choose an appropriate set of control decisions $\CD$,
  and a fusion rule $f$,
  and synthesize a set $\nset$ of supervisors,
  such that $L(f_\nset / G) = L(E)$.
\end{prob}

We usually study the condition for a class of DSCOP
for $\CD$ and $f$ that are fixed \emph{a priori}.
See also \cref{rmk:cd-f}.
In particular, $\CD$ and $f$ should be independent of any specific $G$ and $E$.
In practice, one may choose whatever $\CD$ and $f$ necessary to solve the problem at hand.
Fixing $\CD$ and $f$
allows us to classify pairs of $G$ and $E$ according to the $\CD$ and $f$ sufficient for the decentralized control problem to be solvable,
and thus allows comparison among pairs of $\CD$ and $f$.

\subsection{Epistemic Logic}

\citet{Ricker2000, Ricker2007} observed that
reasoning about the decision\-/making
of decentralized supervisors could be facilitated
using formal reasoning about knowledge,
via epistemic logic.
Although formal conditions for solving DSCOP can be described
using conditions on strings in languages,
and hence do not require a formal logic description,
epistemic logic provides a natural modelling paradigm
that parallels natural languages,
thus giving better intuition into the reasoning
behind the decisions that supervisors make.
Specifically,
the epistemic operator in the language expresses concepts such as
``agent $i$ \emph{knows} that a certain event must be disabled''.
Imbuing supervisors with such anthropomorphic capabilities
sets the stage for decision\-/making to be linked to
``knowledge'' that the agents Have
\--- much as humans base their decision\-/making on
what they know or don't know about a certain situation.
The work in \citet{Ricker2000, Ricker2007} uses epistemic logic
as a way to speak about what knowledge supervisors
must possess for a problem to be solvable,
but it does not capitalize on the link between knowledge and action
to relate a supervisor's decision\-/making directly to its knowledge
in an immediately apparent way.
% which is what we will show in this paper.

Epistemic logic as used in distributed computing problems
was first presented by \citet{Halpern1990}.
See \citet{Fagin2004} for more details.
We provide in the remainder of this section the concepts
from epistemic logic needed to understand our work.

\begin{defn}
  For a fixed set $\vset$ of variables, where $v$ denotes some element of $\vset$, and 
  a fixed finite set $\nset$ of agents, where $i$ denotes some element of $\nset$,
  the set of epistemic modal formulae is defined inductively by the following grammar:
  \begin{grammar}
    %Epistemic Modal Formulae
    & $S, T$
    & \bnfas &
    $
    (v)
    \bnfaltBRKn{propositional variable $v$}
    (\lnot S)
    \bnfaltBRKn{negation of $S$}
    (S \land T)
    \bnfaltBRKn{conjunction of $S$, $T$}
    (K_i S)
    \BRKn{agent $i$ knows $S$}
    $
  \end{grammar}
  
  % We call $\lnot$, $\land$ propositional operators,
  % and call $K_i$ modal operators, or more specifically, epistemic operators.
\end{defn}

\begin{defn}
 \label{defn:connectives}
  It is conventional to define other connectives from the primitive ones above:
  \begin{itemize}
    \item $(\alpha \lor \beta) =_{df} \lnot(\lnot\alpha \land \lnot\beta)$,
    \item $(\alpha \implies \beta) =_{df} (\lnot\alpha \lor \beta)$
  \end{itemize}  
\end{defn}

Where convenient, we use the connectives defined above to express ideas,
but when reasoning about epistemic formulae,
we assume that the connectives of Definition~\ref{defn:connectives} have all been syntactically expanded,
so that we only have to deal with primitive ones.

We omit parentheses according to the following precedence convention:
unary operators $\lnot, K_i$ bind tightest, then $\land, \lor, \implies$.

The semantics of epistemic formulae are given through the use of a structure called a Kripke structure.

\begin{defn}
  For some $\vset$ and $\nset$,
  a Kripke structure, or simply a frame $I$ is
  \[(W, \pi, \setidx{\sim_i}{i \in \nset})\]
  where
  \begin{itemize}
    \item $W$ is a finite set of possible worlds, or states%
    \footnote{
      The term ``states'' should cause no confusion in this context,
      since the worlds in the frames we construct in this work
      happen to be states of some FSA.
    }.
    \item $\pi : W \times \vset \to \set{\true, \false}$
      evaluates each propositional variable in $\vset$ at each possible world in $W$ to either $\true$, or $\false$.
    \item For each $i \in \nset$, ${\sim_i} \subseteq W \times W$ is the accessibility relation over possible worlds,
      and we say world $w'$ is considered by agent $i$ as an epistemic alternative if $w' \sim_i w$.
  \end{itemize}
\end{defn}

Whereas the accessibility relations are commonly required to be equivalence relations over $W$,
the formal construction we will present uses relations that are not reflexive,
and are thus partial equivalence relations.
Hence we denote accessibility relations as $\sim$,
and reserve $\simeq$ for discussions in which the relations are, indeed, equivalence relations. Since \cite{Ricker2007} does not distinguish these cases,
they used $\sim$ for the latter.

In our formalism, the propositional connectives (the second and third items in Definition~\ref{defn:semantics} below) are to be understood as usual.
The semantics of the epistemic operator (the last item in Definition~\ref{defn:semantics}) reflect that,
upon observing a sequence of events generated by the plant,
a supervisor can only \textit{know} something (i.e., be certain that it is true), if it is always true
after any sequence (generated by the plant) that looks the same to the supervisor as the sequence of events it has observed.

To reflect the discussion above,
we thus adopt the following formal definition of the semantics of epistemic formulae
as the relation $\models$ of pairs of Kripke structures and worlds,
and epistemic modal formulae,
given inductively over the structure of the formulae.

\begin{defn}
\label{defn:semantics}
  \begin{itemize}
    \item $(I, w) \models v$ iff $\pi(w, v) = \true$
    \item $(I, w) \models \lnot S$ iff it is not the case that $(I, w) \models S$
    \item $(I, w) \models S \land T$ iff $(I, w) \models S$ and $(I, w) \models T$
    \item $(I, w) \models K_i S$ iff for all $w' \in W$ such that $w' \sim_i w$, $(I, w') \models S$.
  \end{itemize}
\end{defn}

In our discussions, it will often be the case that many epistemic expressions are evaluated against the same pair of $I, w$; in such cases, for simplicity, we will write $S$ in place of $(I, w) \models S$.

% \footnote{
%   Readers familiar with modal logics should note that by stating ``have $S$''
%   we do not mean that $S$ is valid,
%   i.e., $S$ evaluates to $\true$ at every interpretation and every world,
%   since that concept has not been defined here.
% }.

\subsection{Inference\-/Observability}

We can now demonstrate how the epistemic approach by \citet{Ricker2007}
can be adapted and exploited to describe the architecture by \citet{Yoo2004}.
Our approach involves 
\begin{enumerate*}
  \item accessibility relations that differ from those used by \citet{Ricker2007}, and
  \item a deliberate line-by-line correspondence (which we call ``coupling'') between
    the expression of the solvability condition
    and the description of the supervisors.
\end{enumerate*}  The notation and concepts in the first part of this section are adopted from \citet{Ean2021a}.

Consider a plant $G$, with legal behaviour prescribed by
a subautomaton $E$ of $G$, and
$n$ pairs of sets of controllable/observable events. 
For each supervisor $i$ ($i \in \nset$), we construct 
$G^{obs}_i = P_i(G)$, which is effectively an observer of the plant from supervisor $i$'s point of view. The state set, $Q^{obs}_i$, of each observer is $\setcomp{\Sigma_{i,uo}\text{-closure of } q}{q \in Q}$.
% Construct $G^{obs}_i = P_i(G)$ for each $i$,
% where it can also be interpreted $Q^{obs}_i = \setcomp{\Sigma_{i,uo}\text{-closure of } q}{q \in Q}$.
In other words, the supervisor cannot distinguish $G$ and $G^{obs}_i$ by only observing sequences of events generated by these two FSA.

Next we construct a composite structure that will allow us to keep track of plant behaviour and each supervisor's view of the corresponding plant behaviour. We do this through the construction
 $G' = G \times G^{obs}_1 \times \dots \times G^{obs}_n = (\Sigma, Q', \delta', q_0')$,
where $Q' \subseteq Q \times Q^{obs}_1 \times \dots \times Q^{obs}_n \subseteq Q \times \powerset{Q} \times \dots \times \powerset{Q}$,
$\delta'$ is component-wise application of $\delta$ and $\delta^{obs}_i$ for $i \in \nset$,
$q_0' = (q_0, q^{obs}_{0, 1}, \dots, q^{obs}_{0, n})$ where $q^{obs}_{0, i} \in Q^{obs}_{0, i}$ and thus $q^{obs}_{0, i} \subseteq Q$ for $i \in \nset$.

Our composite structure 
$G'$ generates the same language as $G$ does, however the Cartesian product of states forming $Q'$ allows us to track more information than that available by simply tracking the sequence of states in $Q$ visited by some sequence of events in the plant language.  Namely, $(q, q^{obs}_1, \dots, q^{obs}_n) \in Q$ records not only the current state $q$ of $G$,
but also each supervisor's estimate $q^{obs}_i$ of 
the set of states the plant could possibly be in based on supervisor $i$'s observation, for each $i \in \nset$. For the accessible part of $G'$, we always have that $q \in q^{obs}_i$ for all $i \in \nset$, which is to be expected since the actual plant state should always be a state that any observer \textit{thinks} the plant could be in.

Although the automaton $G'$ is not necessarily isomorphic to $G$, its behaviour is identical to that of $G$, namely, 
it is always the case that $L(G') = L(G)$. In other words, from a behavioural standpoint, $G$ and $G'$ cannot be distinguished by observations of generated events.  Moreover, while $G^{obs}_i$ and $G$ are not distinguishable by the particular supervisor $i$,
$G'$ and $G$ are not distinguishable by \textit{any} observer (even one observing $\Sigma$). Consequently, if $G$ specifies a plant, one can think of that plant as also being modelled by $G'$.
Automaton $G'$ can be computed off-line,
and therefore its information is available to all supervisors.

Now we are ready to construct the Kripke structure
against which the expression of inference\-/observability
is interpreted.

We will start by letting $W = Q'$. To avoid multiple arguments with both subscripts and superscripts, we will write $(w_e, w_1, \dots, w_n)$ for an element of $W$ instead of $(q, q^{obs}_1, \dots, q^{obs}_n)$.

Next we construct accessibility relations $\simeq_i$ such that $w \simeq_i w'$
whenever $w_i = w'_i$,
as was done in the work of \citet{Ricker2007}.
Since the accessibility relations $\simeq_i$ are clearly equivalence relations, we can denote $\setcomp{w' \in W}{w' \simeq_i w}$ by $[w]_{\simeq_i}$,
or simply $[w]_i$.

Again, in a similar way as was done in \citet{Ricker2007}, we create propositional variables and their evaluation to capture
when an event $\sigma$ is possible in the plant ($\sigma_G$)
or when it is legal ($\sigma_E$).
Hence we create the set of propositions
$\vset = \Union_{\sigma \in \Sigma} \set{\sigma_G, \sigma_E}$,
and define their evaluations as
\[
  \begin{aligned}
    \pi(w, \sigma_G) &=
      \begin{cases} 
        \true  & \delta^G(\sigma, w)! \\
        \false & \text{otherwise}
      \end{cases}
    \\
    \pi(w, \sigma_E) &=
      \begin{cases} 
        \true  & \delta^E(\sigma, w)! \\
        \false & \text{otherwise}
      \end{cases}
  \end{aligned}
\]
The intended meaning of  $\pi(w, \sigma_G) = \true$  is that $\sigma$ can physically occur at state $w$,
as specified by $G$;
whereas $\pi(w, \sigma_E) = \true$ indicates that $\sigma$ is legal and should be allowed to happen.
It follows that $\pi(w, \sigma_E) = \true \implies \pi(w, \sigma_G) = \true$,
which reflects the fact that $E$ is a subautomaton of $G$.

With the arguments all defined, we can now let the Kripke structure be
$\overline I = (W, \pi, \setidx{\simeq_i}{i \in \nset})$. We denote a Kripke structure as $\overline I$ whenever it is constructed with accessibility relations $\simeq_i$ that are equivalence relations, so that we can reserve the notation $I$ (without the overline) for a construction we will discuss in the next section (where the accessibility relations are not equivalence relations).

Although
the constructed Kripke structure $\overline I$
is parameterized over specific $G$, $\setidx{P_i}{i \in \nset}$,
and $E$, since in our discussions we will not need to  simultaneously consider multiple sets of these entities,
but assume an indefinite one, we write simply $\overline I$,
rather than $\overline I(G, P_1, \cdots, P_n, E)$.

The notion of inference\-/observability was previously defined by
\citet{Ricker2007} as follows.

\begin{defn}
  \label{defn:inf-obs}

  $\overline I$ (or $E$) is said to be inference-observable whenever
  for all $\sigma \in \Sigma$, for all $w \in W$,
  \begin{subequations}
    \renewcommand{\theequation}{\theparentequation.\arabic{equation}}
    \begin{alignat}{3}
      \Exists{i, j \in \nset_\sigma} (\overline I, w) \models
      &      && K_i(\sigma_E \implies K_j (\lnot \sigma_G \lor \sigma_E))
      \label{eq:inf-obs-lr-1}
      \\
      & \lor && \lnot \sigma_G \lor \sigma_E
      \label{eq:inf-obs-lr-2}
    \end{alignat}
  \end{subequations}
\end{defn}

Roughly speaking,
the expression of inference\-/observability can be read in the following way.
At every state $w$, either at least one supervisor $i$ can unambiguously make a control decision (\cref{eq:inf-obs-lr-1}),
or if none of the supervisors can make a control decision,
then it must be the case that event $\sigma$ can be enabled (\cref{eq:inf-obs-lr-2}),
hence the fused decision will be the default $\fde$.
The expression \cref{eq:inf-obs-lr-1}, however, 
does not tell us what decision supervisor $i$ would issue,
or why it issues such a decision.
We will provide in \cref{thm:inf-obs-split} an alternative expression that tells us which supervisor issues which decision.

\citet{Ricker2007} show that inference\-/observability is the necessary and sufficient condition
to solve DSCOP when the supervisors are allowed to infer the knowledge of other supervisors
to some extent.
In \cite{Ricker2007} there are four control decisions,
where the fusion rule is recalled as in \cref{tbl:ricker_fusion}.
Informally our interpretation is as follows.
When the control decision $\cdon$ (resp., $\cdoff$) is present,
the fused decision is guaranteed to be $\fde$ (resp., $\fdd$),
which is why $\cdon$ and $\cdoff$ have been traditionally denoted simply by $\fde$ and $\fdd$.
Based on this fact,
a supervisor issues $\cdon$ (resp., $\cdoff$) for an event $\sigma$
when the supervisor is unambiguously certain
(or in the epistemic logic idiom, ``the supervisor knows'')
that the plant is in a state at which the event $\sigma$ can be enabled (resp., disabled).
This situation is described by the epistemic formula
$K_i(\lnot \sigma_G \lor \sigma_E)$ (resp., $K_i(\lnot \sigma_E)$).

In addition,
the fusion rule allows supervisors to infer the knowledge of other supervisors.
A supervisor $i$ may be uncertain whether an event $\sigma$ can be enabled or disabled,
but it could be the case that among all possible states the plant could be at (all possible worlds) as seen by supervisor $i$,
either the event $\sigma$ can be disabled,
or otherwise if $\sigma$ must be enabled (expressed as $\sigma_E$),
then some \emph{other} supervisor $j$ knows that $\sigma$ can be enabled (expressed as $K_j(\lnot \sigma_G \lor \sigma_E)$),
hence supervisor $j$ will thus issue $\cdon$.
In such a situation,
supervisor $i$ can bet on disabling $\sigma$,
and let supervisor $j$ correct the decision if that bet is a mistake.
This situation is thus described as $K_i(\sigma_E \implies K_j(\lnot \sigma_G \lor \sigma_E))$.
To allow supervisor $j$'s $\cdon$ decision to prevail if the plant is in a state at which $\sigma$ must be enabled,
supervisor $i$ cannot issue $\cdoff$,
since otherwise the fused decision would then be undefined,
and it would be nonsense to define the fused decision for such a case.
Hence a weaker form of the decision, $\cdwoff$,
is intended for such a situation.
The decision $\cdwoff$ was called ``\textbf{conditional off}''
in \citet{Yoo2004} and in \citet{Ricker2007}.

Finally,
the decision $\cdabs$ is used when none of the cases above prevail,
so that a supervisor abstains from voting.
This decision was called ``\textbf{don't know}''
but for reasons we will see in \cref{sec:complete},
it is better to denote this decision as ``$\cdabs$''.
Because it is possible that all supervisors abstain,
to make the fused decision defined,
\citet{Ricker2007} chose to, so to speak, default the decision to $\fde$.

This result is stated formally as follows.

\begin{table}
  \centering
  \begin{tabular}{c c | c}
    $cd_i$      & $cd_j$      & $f_\sigma(\set{cd_i, cd_j})$ \\
    \hline
    $\cdon$     & $\cdon$     & $\fde$ \\
    $\cdon$     & $\cdwoff$   & $\fde$ \\
    $\cdon$     & $\cdabs$    & $\fde$ \\
    \hline
    $\cdoff$    & $\cdoff$    & $\fdd$ \\
    $\cdoff$    & $\cdon$     & $\fdd$ \\
    $\cdoff$    & $\cdabs$    & $\fdd$ \\
    \hline
    $\cdwoff$   & $\cdoff$    & $\fdd$ \\
    $\cdwoff$   & $\cdwoff$   & $\fdd$ \\
    $\cdwoff$   & $\cdabs$    & $\fdd$ \\
    \hline
    $\cdabs$    & $\cdabs$    & $\fde$ \\
  \end{tabular}
  \caption{Fusion rule used by \citet{Ricker2007} and \cref{prop:inf-obs}, where $i, j \in \{1, 2\}$ and $i \ne j$.}
  \label{tbl:ricker_fusion}
\end{table}

\begin{prop}[\cite{Ricker2007}]
  \label{prop:inf-obs}
  With a set of control decisions $\CD = \set{\cdon, \cdoff, \cdwoff, \cdabs}$,
  a fusion rule $f$ defined as in \cref{tbl:ricker_fusion},
  two supervisors $(G^{obs}_i, \KP_i)$,
  where $\KP_i : Q^{obs}_i \times \Sigma \to \CD$ as defined in \cref{fig:ricker_KP},
  solve the DSCOP
  iff
  $\overline I$ is inference-observable.\hfill\propSymbol

  % TODO not use figure environment
  \begin{figure*}[htbp]
    \centering

    \[
      \KP_i(w, \sigma) =
      \begin{cases}
          \cdon
          & (\overline I, w) \models
          % \\
          % &\quad
            \begin{alignedat}[t]{3}
              &  \phantom{{}\land} % HACK!
              && \phantom{{}\lnot} % HACK!
              && K_i(\lnot \sigma_G \lor \sigma_E)
            \end{alignedat}
          \\
          \cdoff
          & (\overline I, w) \models
          % \\
          % &\quad
            \begin{alignedat}[t]{3}
              &       && \lnot && K_i(\lnot \sigma_G \lor \sigma_E) \\
              & \land &&       && K_i(\lnot \sigma_E)
            \end{alignedat}
          \\
          \cdwoff
          & (\overline I, w) \models
          % \\
          % &\quad
            \begin{alignedat}[t]{3}
              &       && \lnot && K_i(\lnot \sigma_G \lor \sigma_E) \\
              & \land && \lnot && K_i(\lnot \sigma_E) \\
              & \land &&       && K_i(\sigma_E \implies K_j(\lnot \sigma_G \lor \sigma_E)) \\
              &       &&       && \text{for $j \in \nset_\sigma$} %{for $j \ne i$} \; \text{\textsuperscript{note}}
            \end{alignedat}
          \\
          \cdabs
          & \text{otherwise}
      \end{cases}
    \]
    
    % {\small
    % Note:
    % \citet{Ricker2007} did not explicitly state the condition $j \ne i$.
    % % But if $j = i$, either $K_i(\lnot \sigma_E)$ or $K_i(\sigma_E \implies K_j(\lnot \sigma_G \lor \sigma_E))$ will be true.
    % In the presence of \cref{lem:split-big},
    % it is superficial,
    % but we nonetheless state it for explicitness.
    % }

    \caption{The knowledge-based control policy used by \cref{prop:inf-obs}}
    \label{fig:ricker_KP}
  \end{figure*}\NoEndMark

\end{prop}

The statement of \cref{prop:inf-obs},
however, requires some attention.
First, while not stated explicitly,
the construction of the fusion rule and the proof of this proposition
give the impression that the work of \citet{Ricker2007} assumes all events are controllable.

Also,
on the one hand, as indicated by their proof,
by ``solving the DSCOP'',
\citet{Ricker2007} implicitly meant that
for all $\sigma \in \Sigma$, 
for \emph{all} $s \in L(G)$,
with letting $w = \delta'(s, q_0')$ (such $w$ exists since $s \in L(G)$),
if $(I, w) \models \sigma_G \land \lnot \sigma_E$
(i.e., $\sigma$ has to be disabled after $s$),
then it is required that $f_\nset(s, \sigma) = \fdd$.
In particular, the above has to hold even for $s \in L(G), s \not \in L(E)$.
On the other hand,
the condition of inference\-/observability quantifies over all worlds,
including those reachable by only illegal strings.
This is illustrated by the following example.

\begin{ex}
  Consider the following example:
  The plant depicted in \cref{fig:bad-G},
  where double circled states are those in $Q^E$,
  and $\Sigma = \set{\alpha, \gamma}$.
  Consider the case of one supervisor
  (since \cref{prop:inf-obs} is claimed to hold for two supervisors \cite{Ricker2007},
  one may simply take as the second supervisor an identical copy of the first one)
  where $\Sigma_o = \emptyset$ and $\Sigma_c = \set{\gamma}$.

  Following the construction defined in this section,
  \cref{fig:bad-G^obs} depicts the supervisor's perception of the plant.
  \cref{fig:bad-G'} depicts jointly the plant states and the supervisors' mind states,
  where the accessibility relations are not explicitly drawn but can be deduced.

  \begin{figure*}[t!]
    \begin{subfigure}[t]{0.30\textwidth}
        \centering
        \includegraphics[width=0.8\textwidth]{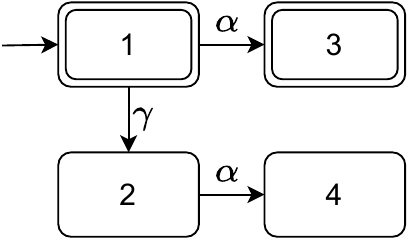}
        \caption{The plant $G$}
        \label{fig:bad-G}
    \end{subfigure}%
    ~
    \begin{subfigure}[t]{0.30\textwidth}
        \centering
        
        \sbox0{\includegraphics[width=0.8\textwidth]{./revising-reachability/G.pdf}}
        \sbox1{\includegraphics[]{./revising-reachability/G.pdf}}
        \makeatletter
        \Gscale@div\myscale{\wd0}{\wd1}
        \makeatother

        \raisebox{\dimexpr.5\ht0-.5\height}{%
          \includegraphics[scale=\myscale]{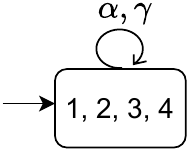}
        }
        \caption{The supervisor's observation $G^{obs} = P(G)$}
        \label{fig:bad-G^obs}
    \end{subfigure}%
    ~
    \begin{subfigure}[t]{0.30\textwidth}
        \centering
        \includegraphics[width=0.9\textwidth]{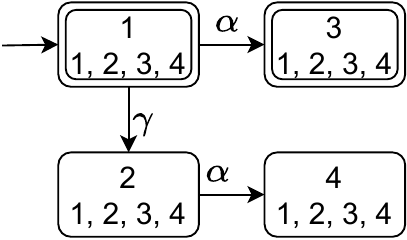}
        \caption{
          The automaton $G' = G \times G^{obs}$.
          A state $(q_G, q^{obs})$ is
          represented in the figure with $q$, $q^{obs}$ stacked vertically in that order.
        }
        \label{fig:bad-G'}
    \end{subfigure}
    \caption{A language that is not inference-observable but for which control exists.}
  \end{figure*}

  Clearly $L(E)$ is not inference-observable in the sense of \cref{defn:inf-obs} (or \cite{Ricker2007}),
  since $(\overline I, (1, \set{1,2,3,4})) \models \gamma_E$
  but $(\overline I, (2, \set{1,2,3,4})) \models \lnot \gamma_E$
  and hence inference\-/observability does not hold at, in particular, world $(1, \set{1,2,3,4})$.

  However $L(E)$ can still be synthesized by disabling $\gamma$ and enabling $\alpha$
  at state $1$.
  Since $\gamma$ is disabled, the plant will not transit to state $2$ and hence $\alpha$
  does not have to be disabled at that state.
  In other words, requiring that certain propositions hold at worlds
  that will not survive control is an unnecessarily strong requirement.
\end{ex}

This example also demonstrates how depicting $G'$ instead of $\overline I$ assists our intuitive understanding
of the situation.

While both $G'$ and $\overline I$ have as their nodes $Q'$,
$\overline I$, unlike $G'$,
illustrates only the accessibility relations but loses the transition relation.
On the other hand,
although not explicitly depicted, 
the accessibility relations can be deduced from $G'$ (to be specific, $Q'$).
Hence, we feel that there is no particular need to explicitly depict the accessibility relations as edges.

\section{Extending inference\-/observability}

We realized,
with the example at the end of the previous section being one of the indications,
that adopting the modal approach of \citet{Ricker2007} to the research of
decentralized supervisory control in general
is not as trivial an extension as what we had anticipated.
Specifically, what our tasks amount to is as follows.
To address the example at the end of the previous section,
\cref{sec:acc-rel} proposes, as one of the many equally good alternatives,
a revision of the accessibility relations.
While the content of \cref{sec:control} is not essential for the work of \citet{Ricker2007} itself,
we found that a slight adjustment to the expression of inference\-/observability
to separate out the condition of what has traditionally been known as \emph{controllability}
will otherwise be beneficial when other studies of decentralized problems are to be cast in the modal approach.
Then \cref{sec:split} further restates inference\-/observability
to create what we call \emph{coupling} between the solvability condition
and the decision policy.
\cref{sec:multi} extends from two supervisors to arbitrarily many supervisors.
Finally, we demonstrate,
with the specific example of the generalized inference\-/observability \citep{Yoo2004},
our claim that the modal approach can be used as a universal formalism in the study of decentralized supervisory control.

\subsection{Accessibility Relation}
\label{sec:acc-rel}

As we have seen at the end of the previous section,
the current expression of the inference\-/observability condition \citep{Ricker2007}
requires that the correct control decisions must be made
even after illegal sequences.
This is unnecessary:
if control decisions can always be made so the system never generates an illegal sequence,
we'd have no obligation at all to make any decision after an illegal sequence.
Equivalently,
since legal behaviours are prefix-closed,
once an illegal sequence is generated,
there is no way we can steer the plant back to legal behaviour.

A naive approach to mend inference\-/observability is as follows:
one may exploit prefix-closedness of the language $L(E)$
and introduce an additional atomic proposition $w_E$ to express that the world $w$ is reachable by legal sequences,
then amend all modal sub-expressions $K_i(\phi)$ to $K_i(w_E \implies \phi)$
and finally replace the entire expression $\phi$ with $w_E \implies \phi$.
This approach, while leading to a better condition,
greatly complicates the matter.
If we take this approach,
with $w_E \implies \cdot$ floating ubiquitously,
the expression would be too complex for an informal reading,
and hence would convey little conceptual intuition,
and thus defy the purpose of expressing ideas in epistemic logic.

The approach we propose here,
in contrast,
strives to 
preserve the form of the earlier work of DSCOP by \citet{Ricker2007}.
As much as possible,
so any previous understanding would easily extend to the work here.
As we shall see,
there is a solution such that the epistemic expression stays intact,
with only a small modification to the Kripke structure
(to be precise, the accessibility relations) needed.

We construct the interpreted system (the Kripke structure) exactly in the same way,
except that we
construct accessibility relations $\sim_i$ such that $w \sim_i w'$
whenever $w_e \in Q^E \land w'_e \in Q^E \land w_i = w'_i$.
Particularly note that $\sim_i$ is an equivalence relation on $\setcomp{w \in W}{w_e \in Q^E}$,
and for all $w$ such that $w_e \not\in Q^E$,
$w$ has no referent nor relatum (participating $\sim_i$).
Hence, the relations $\sim_i$ are \emph{partial equivalence relations}.
It is reasonable to consider the equivalence class
$[w]_{\sim_i}$, or simply, $[w]_i$,
for $w_e \in Q^E$.
Only with an abuse of notation,
let $[w]_i = \emptyset$ for $w_e \not \in Q^E$.
Informally,
one may interpret $[w]_i$ as containing exactly the worlds that are epistemic alternatives to $w$ as perceived by supervisor $i$.
One can observe that ${\sim_i} \subseteq {\simeq_i}$.

To signify the difference,
we denote any frame constructed with the original equivalence accessibility relations as $\overline I$,
and any frame constructed with our new accessibility relations as simply $I$.

One may object to the use of non-equivalence relations as accessibility relations in an epistemic frame,
especially when they are not even reflexive,
which contradicts what one might
understand knowledge to be \citep{Fagin2004}. % TODO does not read well
To relieve this worry,
we give a number of arguments.

Informally,
when a supervisor looks for epistemic alternatives,
it no longer consider illegal states as possible,
as the supervisor can be certain,
that as long as control decisions have been made correctly along the way,
the system is certainly not in an illegal state.
On the other hand, none of the states could be an epistemic alternative to one that is illegal,
including the illegal state itself.
Hence being in an illegal state would be considered an absurdity by a supervisor,
and one can show,
as an epistemic analogy to the EFQ (ex falso quodlibet),
that $(I, w) \models K_i(\phi)$ for arbitrary $\phi$ at any illegal state $w$ (one such that $w_E \not \in Q^E$).
In this way, we encapsulate (or pack) the ubiquitous $w_E \implies \cdot$ appearing in the naive approach inside the modal operators.

As the change only occurs within the Kripke structure,
the rest of the inference\-/observability condition can be expressed exactly as it has been in \cref{defn:inf-obs} when only two supervisors are to be constructed.
Hence we have the following updated definition of inference\-/observability.

\begin{defn}
  \label{defn:inf-obs-corrected}

  $I$ (or $E$) is said to be inference-observable whenever
  for all $\sigma \in \Sigma_c$, for all $w \in W$ such that $w_e \in Q^E$,
  \begin{equation}
    \begin{alignedat}{3}
      \label{eq:inf-obs-corrected}
      \Exists{i, j \in \nset_\sigma} (I, w) \models
      &      && K_i(\sigma_E \implies K_j (\lnot \sigma_G \lor \sigma_E)) \\
      & \lor && \lnot \sigma_G \lor \sigma_E
    \end{alignedat}
  \end{equation}
\end{defn}

\subsection{Separating Controllability}
\label{sec:control}

By intention, we have made
\cref{defn:inf-obs-corrected} differ from that of \cref{defn:inf-obs}
in yet one more way: $\sigma$ is quantified over $\Sigma_c$, instead of $\Sigma$.
This difference deserves some explanation.

Readers more familiar with traditional works
in supervisory control problems with partial observations,
with centralized \citep{Lin1988b} and decentralized \citep{Rudie1992} approaches,
will notice,
such works usually propose two orthogonal conditions for the problem to be solvable.
In the decentralized case \citep{Rudie1992},
one of the conditions, co\-/observability, captures the same idea as inference\-/observability,
namely, does the plant, at any point,
appear unambiguous to at least one supervisor,
where being unambiguous is in the sense
that the same control decision is appropriate in any state
a supervisor thinks the plant could reasonably be in.
The other condition, controllability,
considers whether the desired behaviour is at least implementable
by a centralized, monolithic supervisor,
which observes every event observable to any of the decentralized supervisor,
and controls all controllable events.
It is thus reasonable for such readers to ponder over
the absence of controllability in the work of \citet{Ricker2007}.

In fact, controllability can be incorporated in the corrected inference\-/observability
as \citet{Ricker2007} did without stating so.
However, later as we make extensions to inference\-/observability,
doing so will be quite awkward and the result would obscure, rather than assist, our understanding.
Moreover, as we have demonstrated in our earlier work \citep{Ean2021a},
the fact that controllability and an architecture's observability condition
can be combined into a single expression is only a happy coincidence
for only some architectures.
Hence we choose not to do so.
We still provide the following lemma to bridge the work of \citet{Ricker2007} to this work.

\begin{thm}
  Inference\-/observability, as defined in \cref{defn:inf-obs-corrected}
  (which we write in condensed form using ellipses),
  \[
    \begin{alignedat}[t]{3}
      & \All{\sigma \in \Sigma_c} \All{w \in Q' \text{ such that } w_e \in Q^E}
        \\ &\quad
        \begin{alignedat}{3}
          (I, w) \models
            &      && \dots \\
            & \lor && \lnot \sigma_G \lor \sigma_E
        \end{alignedat}
    \end{alignedat}
  \]
  together with
  \[
    L(E)\Sigma_{uc} \sect L(G) \subseteq L(E)
  \]
  i.e., the prefixed-language $L(E)$ is controllable,
  is equivalent to
  \[
    \begin{alignedat}[t]{3}
      & \All{\sigma \in \Sigma} \All{w \in Q' \text{ such that } w_e \in Q^E}
        \\ &\quad
        \begin{alignedat}{3}
          (I, w) \models
            &      && \dots \\
            & \lor && \lnot \sigma_G \lor \sigma_E
        \end{alignedat}
    \end{alignedat}
  \]
\end{thm}

The theorem can be proven in a manner
analogous to what is demonstrated
in our previous work \citep{Ean2021a}.

\begin{proof}
  To begin, we first rewrite controllability of $L(E)$ as
  \[
    \All{\sigma \in \Sigma_{uc}}
    s \in L(E) \land s\sigma \in L(G) \implies s\sigma \in L(E)
  \]
  which is equivalent to
  \[
    \All{\sigma \in \Sigma_{uc}}
    \begin{aligned}[t]
      & \All{w \in Q' \text{ such that } w_e \in Q^E}
        \\ &\quad
        (I, w) \models
          \lnot \sigma_G \lor \sigma_E
    \end{aligned}
  \]

  Now, inference\-/observability and controllability together is equivalent to
  \[
    \All{\sigma \in \Sigma}
    \begin{alignedat}[t]{3}
      && \sigma \in \Sigma_c \implies &
      \begin{aligned}[t]
        & \All{w \in Q' \text{ such that } w_e \in Q^E}
          \\ &\quad
          \begin{alignedat}{3}
            (I, w) \models
              &      && \dots \\
              & \lor && \lnot \sigma_G \lor \sigma_E
          \end{alignedat}
      \end{aligned}
      \\
      \land
      && \sigma \in \Sigma_{uc} \implies &
      \begin{aligned}[t]
        & \All{w \in Q' \text{ such that } w_e \in Q^E}
          \\ &\quad
          (I, w) \models
            \lnot \sigma_G \lor \sigma_E
      \end{aligned}
    \end{alignedat}
  \]
  
  The part omitted by $\dots$ begins by quantifying over $\nset_\sigma$ existentially,
  which is empty for $\sigma \in \Sigma_{uc}$,
  and hence is trivially false.
  Thus we can attach it to the second conjunct and have equivalently
  \[
    \All{\sigma \in \Sigma}
    \begin{alignedat}[t]{3}
      && \sigma \in \Sigma_c \implies &
      \begin{aligned}[t]
        & \All{w \in Q' \text{ such that } w_e \in Q^E}
          \\ &\quad
          \begin{alignedat}{2}
            (I, w) \models
              &      && \dots \\
              & \lor && \lnot \sigma_G \lor \sigma_E
          \end{alignedat}
      \end{aligned}
      \\
      \land
      && \sigma \in \Sigma_{uc} \implies &
      \begin{aligned}[t]
        & \All{w \in Q' \text{ such that } w_e \in Q^E}
          \\ &\quad
          \begin{alignedat}{2}
            (I, w) \models
              &      && \dots \\
              & \lor && \lnot \sigma_G \lor \sigma_E
          \end{alignedat}
      \end{aligned}
    \end{alignedat}
  \]
  and thus have equivalently
  \[
    \begin{alignedat}[t]{3}
      & \All{\sigma \in \Sigma} \All{w \in Q' \text{ such that } w_e \in Q^E}
        \\ &\quad
        \begin{alignedat}{3}
          (I, w) \models
            &      && \dots \\
            & \lor && \lnot \sigma_G \lor \sigma_E
        \end{alignedat}
    \end{alignedat}
  \]
  as desired.
\end{proof}

\subsection{Splitting Cases}
\label{sec:split}

We take a further step and provide an equivalent expression of inference\-/observability.
The development is analogous to that of \citet[Section 3.6]{Ean2021a}.

\begin{thm}
  \label{thm:inf-obs-split}

  The expression
  \[
    \begin{alignedat}[t]{3}
      \Exists{i, j \in \nset_\sigma} (I, w) \models
      &      && \lnot \sigma_G \lor \sigma_E \\
      & \lor && K_i(\sigma_E \implies K_j (\lnot \sigma_G \lor \sigma_E))
    \end{alignedat}
  \]
  is equivalent to
  \begin{equation}
    \label{eq:inf-obs-split}
    \begin{alignedat}{3}
      \Exists{i, j \in \nset_\sigma, i \ne j} (I, w) \models
      &      && K_i(\lnot \sigma_G \lor \sigma_E) \\
      & \lor && K_i(\lnot \sigma_E) \\
      & \lor && K_i(\sigma_E \implies K_j (\lnot \sigma_G \lor \sigma_E)) \\
      & \lor && \lnot \sigma_G \lor \sigma_E
    \end{alignedat}
  \end{equation}
\end{thm}

Before proving this theorem,
we motivate the need for it.

\begin{rmk}
  \label{rmk:inf-obs-alt}

  \Cref{thm:inf-obs-split}
  applies both to
  inference\-/observability
  and the fusion rule in \cref{fig:ricker_KP} (where $\overline I$ becomes $I$ as discussed).
  When applied to the fusion rule in \cref{fig:ricker_KP},
  \cref{thm:inf-obs-split} says
  that we can add the requirement that $j \ne i$,
  which
  \--- although superfluous \---
  is intuitively helpful.
  When applied to inference\-/observability,
  the alternative expression has a line-by-line correspondence
  with that of the fusion rule.
  We have thus created a \emph{direct coupling} between the problem-solvability
  requirement of inference\-/ inference-observability
  and the control policy to be used when the problem is solvable.
  Not only does it give a better understanding of the inference\-/observability condition than \cref{defn:inf-obs},
  it also yields an easier proof of
  that condition (both the original version of \cite{Ricker2007} and our extended version to be given)
  being the necessary and sufficient condition
  to solve DSCOP
  (using the specific set of control decisions, knowledge-based control policy, and fusion function).
  The proof can be trivially (i.e., mechanically) done by case analysis.
\end{rmk}

% \replaced{
%   To approach a proof of \mbox{\cref{thm:inf-obs-split}},
%   we follow the method analogous to what is presented
%   in \mbox{\citep[Section 3.6.2]{Ean2021a}}.
%   That is, we need the following lemmata.
% }{
  To prove \mbox{\cref{thm:inf-obs-split}},
  we need the following lemmata.
% }

\begin{lem}
  \label{lem:split-small}
  \begin{equation}
    (I, w) \models K_i(\sigma_E \implies K_i(\lnot \sigma_G \lor \sigma_E))
  \end{equation}
  iff
  \begin{equation}
    \begin{alignedat}{3}
      (I, w) \models 
        &      && K_i(\lnot \sigma_G \lor \sigma_E) \\
        & \lor && K_i(\lnot \sigma_E)
    \end{alignedat}
  \end{equation}
\end{lem}

\begin{pf}
  ($\impliedby$):
    We have
    \[
      \begin{aligned}
        & (I, w) \models 
          K_i (\lnot \sigma_E)
        \\
        \implies
        & (I, w) \models
          K_i (\lnot \sigma_E \lor K_i(\lnot \sigma_G \lor \sigma_E))
        \\
        \implies
        & (I, w) \models
          K_i (\sigma_E \implies K_i (\lnot \sigma_G \lor \sigma_E))
      \end{aligned}
    \]
    and
    \[
      \begin{aligned}
        & (I, w) \models
          K_i (\lnot \sigma_G \lor \sigma_E)
        \\
        \implies
        & \All{w' \in [w]_i}
          (I, w') \models 
          K_i (\lnot \sigma_G \lor \sigma_E)
        \\
        \implies
        & \All{w' \in [w]_i} 
          (I, w') \models
          \sigma_E \implies K_i (\lnot \sigma_G \lor \sigma_E)
        \\
        \implies
        & (I, w) \models
          K_i (\sigma_E \implies K_i (\lnot \sigma_G \lor \sigma_E))
      \end{aligned}
    \]
  
  ($\implies$):
  \begin{linewise}
    Assume $(I, w) \models K_i(\sigma_E \implies K_i(\lnot \sigma_G \lor \sigma_E))$.
    Hence equivalently
      $\All{w' \in [w]_i} (I, w') \models \sigma_E \implies K_i(\lnot \sigma_G \lor \sigma_E)$.
    Hence equivalently
      $\All{w' \in [w]_i} (I, w') \models \lnot \sigma_E \lor K_i(\lnot \sigma_G \lor \sigma_E)$. \hfill $(\ast)$
    So we have either
      A: $w_e \in Q^E$; or
      B: $w_e \not\in Q^E$.
    \underline{Case A: $w_e \in Q^E$}
      Hence $[w]_i$ is not empty.
      We have either
        A.1: $\Exists{w' \in [w]_i} (I, w') \models K_i(\lnot \sigma_G \lor \sigma_E)$; or
        A.2: $\lnot \Exists{w' \in [w]_i} (I, w') \models K_i(\lnot \sigma_G \lor \sigma_E)$
        // The trick here is to not split the disjunction in $(\ast)$.
      \underline{Case A.1: $\Exists{w' \in [w]_i} (I, w') \models K_i(\lnot \sigma_G \lor \sigma_E)$}
        Obtain $w'$ such that
          $w' \in [w]_i$ and
          $(I, w') \models K_i(\lnot \sigma_G \lor \sigma_E)$.
        Hence $\All{w'' \in [w']_i} (I, w'') \models K_i(\lnot \sigma_G \lor \sigma_E)$.
        With $[w']_i = [w]_i$,
          we have $\All{w'' \in [w]_i} (I, w'') \models K_i(\lnot \sigma_G \lor \sigma_E)$.
        Hence $(I, w) \models K_i(\lnot \sigma_G \lor \sigma_E)$.
      \underline{Case A.2: $\lnot \Exists{w' \in [w]_i} (I, w') \models K_i(\lnot \sigma_G \lor \sigma_E)$}
        Hence $\All{w' \in [w]_i} (I, w') \models \lnot K_i(\lnot \sigma_G \lor \sigma_E)$.
        With $(\ast)$,
          we have $\All{w' \in [w]_i} (I, w') \models \lnot (\sigma_E)$.
        Thus $(I, w) \models K_i(\lnot \sigma_E)$.
      Together from A.1 and A.2, we have either
        $(I, w) \models K_i(\lnot \sigma_G \lor \sigma_E)$; or
        $(I, w) \models K_i(\lnot \sigma_E)$.
      Thus $(I, w) \models K_i(\lnot \sigma_G \lor \sigma_E) \lor K_i(\lnot \sigma_E)$.
    \underline{Case B: $w_e \not \in Q^E$}
      Hence $[w]_i$ is empty.
      Hence $\All{w' \in [w]_i} (I, w') \models \phi$ holds vacuously true.
      Hence $(I, w) \models K_i(\phi)$ holds vacuously true.
      Hence $(I, w) \models K_i(\lnot \sigma_G \lor \sigma_E)$
        and $(I, w) \models K_i(\lnot \sigma_E)$ hold vacuously true.
      Thus $(I, w) \models K_i(\lnot \sigma_G \lor \sigma_E) \lor K_i(\lnot \sigma_E)$.
    Together from A and B, have
      $(I, w) \models K_i(\lnot \sigma_G \lor \sigma_E) \lor K_i(\lnot \sigma_E)$,
      which is what we wanted.
  \end{linewise}
\end{pf}

\begin{lem}
  \label{lem:split-big}
  \begin{equation}
    \label{eq:split-a}
    \begin{aligned}
      & \Exists{i,j \in \nset_\sigma} \\
      & \quad
        (I, w) \models
          K_i(\sigma_E \implies K_j(\lnot \sigma_G \lor \sigma_E))
    \end{aligned}
  \end{equation}
  iff
  \begin{equation}
    \label{eq:split-b}
    \begin{aligned}
      & \Exists{i,j \in \nset_\sigma, i \ne j} \\
      & \quad
        \begin{alignedat}[t]{3}
          (I, w) \models
            &      && K_i(\lnot \sigma_G \lor \sigma_E) \\
            & \lor && K_i(\lnot \sigma_E) \\
            & \lor && K_i(\sigma_E \implies K_j(\lnot \sigma_G \lor \sigma_E))
        \end{alignedat}
    \end{aligned}
  \end{equation}
\end{lem}

% \begin{pf}[sketch]
%   The result can be proven
%   by considering separately the cases $i = j$ and $i \ne j$,
%   and for the first case, using \cref{lem:split-small}.
% \end{pf}

\begin{pf}
  We show \cref{eq:split-a} is equivalent to \cref{eq:split-b}
  through a sequence of equivalences.
  By splitting cases, we have that \cref{eq:split-a}, i.e.,
  \[
    \Exists{i,j}
      (I, w) \models
      K_i(\sigma_E \implies K_j(\lnot \sigma_G \lor \sigma_E))
  \]
  is equivalent to
  \[
  \begin{aligned}
    &[\Exists{i,j, i = j} (I, w) \models K_i(\sigma_E \implies K_j(\lnot \sigma_G \lor \sigma_E))]
    \\
    \lor
    &[\Exists{i,j, i \ne j} (I, w) \models K_i(\sigma_E \implies K_j(\lnot \sigma_G \lor \sigma_E))]
  \end{aligned}
  \]

  Focusing on the first disjunct, by substituting $j$ with $i$, we have equivalently
  \[
  \begin{aligned}
    &[\Exists{i} (I, w) \models K_i(\sigma_E \implies K_i(\lnot \sigma_G \lor \sigma_E))]
    \\
    \lor
    &[\Exists{i,j, i \ne j} ...]
  \end{aligned}
  \]

  Substituting the first disjunct with the equivalence given by \cref{lem:split-small},
  we have equivalently
  \[
  \begin{aligned}
    & \left[
      \begin{alignedat}{3}
        \Exists{i} (I, w) \models 
          &      && K_i(\lnot \sigma_G \lor \sigma_E) \\
          & \lor && K_i(\lnot \sigma_E)
      \end{alignedat}
      \right]
    \\
    \lor
    &[\Exists{i,j, i \ne j} ...]
  \end{aligned}
  \]

  By quantifying over $j$ in the first disjunct which does not occur free there,
  we have equivalently
  \[
  \begin{aligned}
    & \left[
      \begin{alignedat}{3}
        \Exists{i, j, i \ne j} (I, w) \models 
          &      && K_i(\lnot \sigma_G \lor \sigma_E) \\
          & \lor && K_i(\lnot \sigma_E)
      \end{alignedat}
      \right]
    \\
    \lor
    &[\Exists{i,j, i \ne j} ...]
  \end{aligned}
  \]

  Thus we have equivalently
  \[
    \begin{alignedat}{3}
      \Exists{i, j, i \ne j} (I, w) \models 
        &      && K_i(\lnot \sigma_G \lor \sigma_E) \\
        & \lor && K_i(\lnot \sigma_E) \\
        & \lor && K_i((\sigma_G \land \lnot \sigma_E) \implies K_j(\sigma_E))
    \end{alignedat}
  \]
  which is what we wanted.
\end{pf}

Now we are ready to prove \cref{thm:inf-obs-split}.

\begin{pf}[\cref{thm:inf-obs-split}]
  By \cref{lem:split-big}.
\end{pf}

Finally,
we are ready to restate the significance of inference\-/observability to DSCOP.

\begin{thm}
  \label{thm:inf-obs-corrected}

  With a set of control decisions $\CD = \set{\cdon, \cdoff, \cdwoff, \cdabs}$,
  a fusion rule $f$ defined as in \cref{tbl:ricker_fusion},
  there exists a set $\nset$ of two supervisors
  that solves the DSCOP
  iff
  $I$ is controllable and inference-observable (in the sense of \cref{defn:inf-obs-corrected}).

  Moreover, whenever controllability and inference\-/observability hold,
  the supervisors can be constructed as
  $(G^{obs}_i, \KP_i)$,
  where $\KP_i$ are defined as in \cref{fig:ricker_KP}
  with $\overline I$ replaced by $I$.
\end{thm}

Notice that, aside from using the alternatively expressed inference\-/observability and replacing $\overline I$ by $I$,
this theorem is stated differently than \cref{prop:inf-obs}.
The latter does not assert
that if inference\-/observability fails,
the problem is not solvable by supervisors of some other flavour,
i.e., ones that are not based on the FSA $G^{obs}_i$ and the knowledge-based control policy $\KP$.

To support our claim that epistemic logic can be used as a universal tool
in studies of DSCOP,
we have to make the point
that if the problem is solvable by supervisors of any flavour,
it must always be solvable by supervisors based on the FSA $G^{obs}_i$ and the knowledge-based control policy $\KP$.

Assuming supervisors of a particular flavour can be problematic.
One should however, only assume the fusion rule,
as it suffices to give semantics to the control decisions.
See also \cref{rmk:cd-f}.

This is not seen in traditional approaches to DSCOP such as \cite{Rudie1992}
where $\FD$ is taken to be equal to $\CD$ and the control policy is not
explicitly constructed:
disabling of an event at a state is expressed by the absence of a transition labelled by that event
at the corresponding state of a supervisor, hence the term ``implicit supervisors''.

We do not provide a proof for \cref{thm:inf-obs-corrected}.
A proof could be obtained by altering the proof provided
by \cite{Ricker2007} to \cref{prop:inf-obs} (their Thm. 1) correspondingly.
More significantly, though,
an alternative proof can be easily obtained from the proof of our main result \cref{thm:inf-obs-extended}
due to the line-to-line correspondence between the control policy and inference\-/observability expression.
See also \cref{rmk:inf-obs-alt}.

\subsection{Extending to Arbitrarily Many Supervisors}
\label{sec:multi}

While the framework set up by \cite{Ricker2007} is stated for an arbitrary number of supervisors,
their fusion function is designed for only two supervisors,
and hence also their proof.

One may realize that the expression of inference\-/observability appears to be compatible with arbitrarily many supervisors
because of the existential quantification,
and attempt only a simple extension of the fusion function.

While this approach does yield a broader class of DSCOP to be solvable,
with a weaker\-/than\-/it\-/could\-/be inference\-/observability condition,
it is not as general as possible.

Omitting the fusion function for now,
consider just the (alternatively expressed) inference\-/observability condition
given by \cref{eq:inf-obs-split}.
It is not hard to see that it is equivalent to the expression \cref{eq:inf-obs-multi-bad}:
\begin{subequations}
  \renewcommand{\theequation}{\theparentequation.\arabic{equation}}
  \label{eq:inf-obs-multi-bad}
  \begin{alignat}{3}
    (I, w) \models
    &      && \lOr_{i \in \nset_\sigma} K_i(\lnot \sigma_G \lor \sigma_E)
    \notag \addtocounter{equation}{1} \\
    & \lor && \lOr_{i \in \nset_\sigma} K_i(\lnot \sigma_E)
    \notag \addtocounter{equation}{1} \\
    & \lor && \lOr_{i \in \nset_\sigma} \lOr_{\stackrel{j \in \nset_\sigma}{j \ne i}}K_i(\sigma_E \implies K_j (\lnot \sigma_G \lor \sigma_E))
    \label{eq:inf-obs-multi-bad-3} \\
    & \lor && \lnot \sigma_G \lor \sigma_E
    \notag \addtocounter{equation}{1}
  \end{alignat}
\end{subequations}

Since the number of supervisors is finite,
we rewrote existential quantifiers to n-ary disjunctions,
and distributed then down the expression tree as far as possible.

Compare the condition \cref{eq:inf-obs-multi-bad} above with the following one (\cref{eq:inf-obs-multi}):
\begin{subequations}
  \renewcommand{\theequation}{\theparentequation.\arabic{equation}}
  \label{eq:inf-obs-multi}
  \begin{alignat}{3}
    (I, w) \models
    &      && \dots
    \notag \addtocounter{equation}{2} \\
    & \lor && \lOr_{i \in \nset_\sigma} K_i(\sigma_E \implies \lOr_{\stackrel{j \in \nset_\sigma}{j \ne i}} K_j (\lnot \sigma_G \lor \sigma_E))
    \label{eq:inf-obs-multi-3} \\
    &      && \dots
    \notag
  \end{alignat}
\end{subequations}

These two conditions would be equivalent if $\nset = \{1, 2\}$.
With $j \ne i$, when there are only two supervisors,
once $i$ is fixed, $j$ must also be.
But in general, \cref{eq:inf-obs-multi} is weaker.

Where $j$ is quantified matters:
if we quantify $j$ at where $i$ is quantified as we did in condition \cref{eq:inf-obs-multi-bad-3},
for this line to hold true,
supervisor $i$ has to be able to certify the knowledge of some indefinite but fixed supervisor $j$;
whereas when we do as in \cref{eq:inf-obs-multi-3},
we only require supervisor $i$ to combine the knowledge of a collection of supervisors.
Technically, this distinction is due to the disjunction ($\lOr$)
not commuting across the implicit conjunction hidden in the modal operator $K_i$ ($\forall w' \in [w]_i$).

Since we intend to make further extension of inference\-/observability,
we do not give the corresponding control protocols and fusion rules here.
One will see how to derive them (and proofs of their correctness) from the discussion below.

\subsection{Completing the Decision Set}
\label{sec:complete}

Although the work of \citet{Ricker2007} was based on \citeauthor{Yoo2004}'s work on the use of non-binary control decisions \citep{Yoo2004},
\citeauthor{Ricker2007} did not include the decision $\cdwon$.
Further,
given an a priori partition $\Sigma_c = \Sigma_{c,e} \union \Sigma_{c,d}$,
\citet{Yoo2004} allows one to ``default'' the final decision of an event to either $\fde$ or $\fdd$
when all supervisors are uncertain and issue $\cdabs$.
The construction of \mbox{\citet{Ricker2007}} implicitly assumed that $\Sigma_c = \Sigma_{c,e}$,
so that all events are defaulted to be enabled.
This section adds the capability
of assigning the default decision of some event
to $\fdd$ when all supervisors $\cdabs$.

Hence in this section,
we extend the DSCOP problem to one where
$\CD = \set{\cdon, \cdoff, \cdwon, \cdwoff, \cdabs}$,
i.e., one that includes the complete set of control decisions from \cite{Yoo2004}.
This yields a relaxed condition of inference\-/observability.
We state and prove the result for an arbitrary number of supervisors.
From there,
we will indicate how proofs to claims we made in previous sections can easily be obtained.
Henceforth, unless explicitly stated for comparison,
we use inference\-/observability to refer to the extended condition.

For ease of understanding and compactness,
we define the following shorthand notation for epistemic formulae,
all implicitly parameterized by an event $\sigma$ known from the context.
The notation originated in our previous work \citep[Secion 3.11]{Ean2021a}.

First, phrases regarding the desired decision of $\sigma$:
\begin{alignat*}{3}
  &e
  &&= \lnot \sigma_G \lor \sigma_E
  &&\qquad\text{$\sigma$ can be enabled}
  \\
  &d
  &&= \lnot \sigma_E
  &&\qquad\text{$\sigma$ can be disabled}
  \\
  &\underline{e}
  &&= \sigma_E
  &&\qquad\text{$\sigma$ must be enabled}
  \\
  &\underline{d}
  &&= \sigma_G \land \lnot \sigma_E
  &&\qquad\text{$\sigma$ must be disabled}
\end{alignat*}

We give informal readings respectively:
\begin{itemize}
  \item $\underline{e} = \sigma_E$:
    being equivalent to $\sigma_G \land \sigma_E$,
    can be read as ``$\sigma$ must be enabled to satisfy the control requirement''.
  \item $e = \lnot \sigma_G \lor \sigma_E$: $\sigma$ can be enabled without violating the control requirement.
    Moreover, being equivalent to $\sigma_G \implies \sigma_E$,
    the expression can be read as
    ``if something ought to be decided about $\sigma$, it is enabling, otherwise whatever''. 
  \item $\underline{d} = \sigma_G \land \lnot \sigma_E$:
    can be read as ``$\sigma$ must be disabled to satisfy the control requirement''.
  \item $d = \lnot \sigma_E$:
    equivalent to $\lnot \sigma_G \lor \lnot \sigma_E$,
    and hence equivalent to $\sigma_G \implies \lnot \sigma_E$,
    can be read as
    ``if something ought to be decided about $\sigma$, it is disabling, otherwise whatever''.
    Or more compactly,
    ``$\sigma$ can be disabled without violating the control requirement''.
\end{itemize}

Then, define the modal operator ``someone knows\ldots'':
\[
  S\phi
  =\lor_{i \in \nset_\sigma} K_i\phi
\]

With a supervisor $i$ known from the context,
define a variant of the modal operator ``someone knows''
as ``some other supervisor (other than $i$) knows\ldots'':
\[
  O\phi
  =\lor_{\substack{j \in \nset_\sigma\\j \ne i}} K_i\phi
\]

\begin{defn}
  \label{defn:inf-obs-extended}
  $I$ (or $E$) is said to be inference-observable whenever
  for all $\sigma \in \Sigma_c$,
  there is a certain $\phi_\sigma \in \set{e, d}$ for this $\sigma$,
  so that for all $w \in W$ such that $w_e \in Q^E$,
  we have
  \begingroup
  \allowdisplaybreaks
  \begin{subequations}
    \renewcommand{\theequation}{\theparentequation.\arabic{equation}}
    \label{eq:inf-obs}
    \begin{alignat}{3}
      (I, w) \models
      &      && Se
      \label{eq:inf-obs-1} \\
      & \lor && Sd
      \label{eq:inf-obs-2} \\
      & \lor && S(\underline{e} \implies Oe)
      \label{eq:inf-obs-3} \\
      & \lor && S(\underline{d} \implies Od)
      \label{eq:inf-obs-4} \\
      & \lor && \phi_\sigma
      \label{eq:inf-obs-5}
    \end{alignat}
  \end{subequations}
  \endgroup
\end{defn}

The interpretation of the expression is as follows:
if one of \cref{eq:inf-obs-1,eq:inf-obs-2,eq:inf-obs-3,eq:inf-obs-4} holds,
at least one supervisor unambiguously knows what decision to issue.
In all the worlds such that none of these expressions hold,
i.e., the worlds in which all supervisors abstain,
then either in all such worlds $\sigma$ could be disabled,
or in all such worlds $\sigma$ could be enabled.
This would allow us to ``default'' the decision of an event to either
$\fde$ or $\fdd$.

\Cref{defn:inf-obs-extended} is clearly strictly weaker than \cref{defn:inf-obs-corrected}.

The last ingredient we need before showing that inference\-/observability is necessary and sufficient to solve DSCOP is a characterization of ``solving'' DSCOP.
This characterization is expressed in the following lemma.

\begin{lem}
  A joint supervision $f_\nset$ solves DSCOP 
  iff
  \begin{subequations}
    \renewcommand{\theequation}{\theparentequation.\arabic{equation}}
    \label{eq:solve}
    \begin{align}
      & s \in L(E) \land s\sigma \in L(G) \land \sigma \in \Sigma_{uc}
      \nonumber
      \\ &\quad \implies s\sigma \in L(E)
      \label{eq:solve-1}
      \\
      & s \in L(E) \land s\sigma \in L(G) \land \sigma \in \Sigma_c \land s\sigma \in L(E)
      \nonumber
      \\ &\quad \implies f_\nset(s, \sigma) = \fde
      \label{eq:solve-2}
      \\
      & s \in L(E) \land s\sigma \in L(G) \land \sigma \in \Sigma_c \land s\sigma \not\in L(E)
      \nonumber
      \\ &\quad \implies f_\nset(s, \sigma) = \fdd
      \label{eq:solve-3}
    \end{align}
  \end{subequations}
\end{lem}

The lemma expresses that a solution to DSCOP must ensure that uncontrollable events do not lead to illegality \cref{eq:solve-1},
and that if a controllable event is legal, it is allowed to happen \cref{eq:solve-2},
and if it is illegal, it is prevented from happening \cref{eq:solve-3}.

\begin{pf}
  Directly from the definition of $L(f_\nset / G)$.
\end{pf}

We can now state and prove our main result.

\newcommand{\runonvtt}{\normalfont\fontfamily{cmvtt}\selectfont}
\newcommand{\textvtt}[1]{\text{\runonvtt{#1}}}
\newcommand{\kw}[1]{\text{\textvtt{#1}}}

\newcommand{\ditto}{\ensuremath{''}}

\begin{thm}
  \label{thm:inf-obs-extended}

  With a set of control decisions $\CD = \set{\cdon, \cdoff, \cdwon, \cdwoff, \cdabs}$,
  and for each $\sigma \in \Sigma_c$, a default action $\kw{dft} \in \set{\fde, \fdd}$,
  so that the fusion rule $f_\sigma^{\kw{dft}}$ for $\sigma$ is defined as
  \begin{equation*}
    f_\sigma^{\kw{dft}}(cd) = \left\{
    \begin{array}{llrlrl}
        \fde
      & \text{if } & \cdon  &    \in cd, & \cdoff  &\not\in cd
      \\
        \fdd
      & \text{if } & \cdon  &\not\in cd, & \cdoff  &    \in cd
      \\
        \fde
      & \text{if } & \cdon  &\not\in cd, & \cdoff  &\not\in cd, \\
      &            & \cdwon &\in cd,     & \cdwoff &\not\in cd
      \\
        \fdd
      & \text{if } & \cdon  &\not\in cd, & \cdoff  &\not\in cd, \\
      &            & \cdwon &\not\in cd, & \cdwoff &    \in cd
      \\
        \kw{dft}
      & \text{if } & \cdon  &\not\in cd, & \cdoff  &\not\in cd, \\
      &            & \cdwon &\not\in cd, & \cdwoff &\not\in cd
    \end{array} \right.
  \end{equation*}
  where $cd = \setidx{f_i(P_i(s), \sigma)}{i \in \nset_\sigma}$ for short,
  there exists a set $\nset$ of $n$ supervisors
  that solves the DSCOP
  iff
  $I$ is controllable
  and inference-observable
  (in the sense of \cref{defn:inf-obs-extended}).
\end{thm}

Whenever controllability and inference\-/observability hold,
the construction produced in our proof yields a set of knowledge-based supervisors that solves the DSCOP.

Informally,
for the necessary part of the proof,
we will perform a case analysis on the control decisions a supervisor may issue,
and show that the control requirement cannot be achieved if the language is not inference-observable.

For the sufficient part, 
we will also perform a case analysis.
This part is more complicated, as one cannot simply say ``if line $n$ is true''.
In some situations an event $\sigma$ cannot happen,
and therefore no decision is required,
which justifies that the fusion rule is not defined on some cases.
In other situations, the condition tells not only what fused decision must be achieved,
but also what control decisions each supervisor will make by referring to the control policy,
and it will be immediately apparent that the control decisions issued will be fused to exactly the desired decision.

We now provide our formal proof.

\begin{pf}
  Condition \cref{eq:solve-1} is equivalent to controllability.
  Hence it suffices to show that \cref{eq:solve-2,eq:solve-3} are equivalent to inference\-/observability.

  ($\implies$):
    Suppose there exists such a set $\nset = (f_1, \dots, f_n)$ of $n$ supervisors,
    such that \cref{eq:solve-2,eq:solve-3} hold.

    Suppose, for the sake of contradiction,
    that $L(E)$ is not inference observable.
    I.e., for some $\sigma \in \Sigma_c$,
    for each $\phi_\sigma \in \set{e, d}$,
    there is some $w$ such that $w_e \in Q^E$,
    \begin{subequations}
    \begingroup%
    \allowdisplaybreaks%
      \renewcommand{\theequation}{\theparentequation.\arabic{equation}}
      \begin{alignat}{3}
        (I, w) \models
        &       && \lAnd_{i \in \nset_\sigma} \lnot
          K_i e
        \label{eq:not-inf-obs-1} \\
        & \land && \lAnd_{i \in \nset_\sigma} \lnot
          K_i d
        \label{eq:not-inf-obs-2} \\
        & \land && \lAnd_{i \in \nset_\sigma} \lnot
          K_i (\underline{e} \implies Oe)
        \label{eq:not-inf-obs-3} \\
        & \land && \lAnd_{i \in \nset_\sigma} \lnot
          K_i (\underline{d} \implies Od)
        \label{eq:not-inf-obs-4} \\
        & \land && \lnot \phi_\sigma
        \label{eq:not-inf-obs-5}
      \end{alignat}\endgroup%
    \end{subequations}

    We could proceed by considering either $\phi_\sigma = e$ or $\phi_\sigma = d$,
    since it suffices to derive a contradiction from either of them.
    We choose $\phi_\sigma = e = \lnot \sigma_G \lor \sigma_E$.
    Then by \cref{eq:not-inf-obs-5},
    we have $\sigma_G \land \lnot\sigma_E$ so $\sigma$ must be disabled
    after any sequence leading to state $w$.

    Consider the string $s$ such that
    $\delta'(s, q_0') = w$.
    Such a string must exist and $s \in L(E)$ since $w_e \in Q^E$.
    Now we have $s\sigma \in L(G) - L(E)$,
    hence by \cref{eq:solve-3} it must be that $f_\nset(s, \sigma) = \fdd$.
    By the fusion rule, this can be achieved in two ways:
    either $\Exists{i \in \nset_\sigma} f_i(P_i(s), \sigma) = \cdoff$,
    or alternatively
    $\lnot \Exists{i \in \nset_\sigma} f_i(P_i(s), \sigma) = \cdoff$,
    $\lnot \Exists{i \in \nset_\sigma} f_i(P_i(s), \sigma) = \cdon$ and
    $\Exists{i \in \nset_\sigma} f_i(P_i(s), \sigma) = \cdwoff$.

    If it is the case that $f_i(P_i(s), \sigma) = \cdoff$ for some $i$,
    we can derive a contradiction in the following way:
    from \cref{eq:not-inf-obs-2},
    there is a world $w' \in [w]_i$ and $(I, w) \models \sigma_E$.
    Then $w_e' \in Q^E$ and there is a string $s'$ such that
    $P_i(s') = P_i(s)$, $s'\sigma \in L(E)$.
    Then we have $f_i(P_i(s'), \sigma) = f_i(P_i(s), \sigma) = \cdoff$,
    and by the fusion rule we have $f_\nset(s', \sigma) = \fdd$,
    contradicting the requirement that $f_\nset(s', \sigma) = \fde$.

    If it is the case that
    $\All{i \in \nset_\sigma} f_i(P_i(s), \sigma) \ne \cdoff$,
    $\All{i \in \nset_\sigma} f_i(P_i(s), \sigma) \ne \cdon$ and
    $f_i(P_i(s), \sigma) = \cdwoff$ for some $i \in \nset_\sigma$,
    we can derive contradiction in the following way:
    from \cref{eq:not-inf-obs-3},
    there is a world $w' \in [w]_i$ and $(I, w') \models \sigma_E \land \lAnd_{\stackrel{j \in \nset_\sigma}{j \ne i}} \lnot K_j(\lnot \sigma_G \lor \sigma_E)$.
    In particular, 
    $(I, w') \models \sigma_E$,
    and for all $j \in \nset_\sigma$, $j \ne i$, 
    there is a world $w'' \in [w']_j$
    such that $(I, w'') \models \sigma_G \land \lnot \sigma_E$.
    Since $w_e', w_e'' \in Q^E$,
    there are strings $s', s'' \in L(E)$ such that $\delta'(s', q_0') = w'$ and $\delta'(s'', q_0') = w''$.
    Then we have $f_i(P_i(s'), \sigma) = f_i(P_i(s), \sigma) = \cdwoff$,
    so to enable $\sigma$ after $s'$, i.e., at world $w'$,
    the fusion rule requires $f_j(P_j(s'), \sigma) = \cdon$ for some $j \in \nset_\sigma$, $j \ne i$.
    However there cannot be such a supervisor $j$,
    as otherwise we have $f_j(P_j(s''), \sigma) = f_j(P_j(s'), \sigma) = \cdon$,
    and by the fusion rule we have $f_{\nset_\sigma}(s'', \sigma) = \fde$,
    which violates the control requirement.

    Hence the language must be inference-observable.

  ($\impliedby$):
    Suppose that $L(E)$ is inference-observable.

    We provide a knowledge-based control policy that forms the basis of our solution:
    \begin{equation}
    \begin{aligned}
      & \KP_i(w, \sigma) =
      \\
      & \quad
        \begin{array}\{{llllrlll}. % the dot must be used to denote a null delimiter
            \cdon
          & \rlap{if $(I, w) \models$}\quad
          &       &       & K_i e
          & \land & \lnot & K_i d
          \\
            \cdoff
          & \rlap{if $(I, w) \models$}\quad
          &       &       & \lnot K_i e
          & \land &       & K_i d
          \\
            \cdwon
          & \rlap{if $(I, w) \models$}\quad
          &       &       & \lnot K_i e
          & \land & \lnot & K_i d
          \\
          &
          & \land & \lnot & K_i (\underline{e} \implies Oe)
          & \land &       & K_i (\underline{d} \implies Od)
          \\
            \cdwoff
          & \rlap{if $(I, w) \models$}\quad
          &       &       & \lnot K_i e
          & \land & \lnot & K_i d
          \\
          &
          & \land &       & K_i (\underline{e} \implies Oe)
          & \land & \lnot & K_i (\underline{d} \implies Od)
          \\
            \cdabs
          & \rlap{if $(I, w) \models$}\quad
          &       &       & \lnot K_i e
          & \land & \lnot & K_i d
          \\
          &
          & \land &       & K_i (\underline{e} \implies Oe)
          & \land &       & K_i (\underline{d} \implies Od)
          \\
            \cdabs
          & \multicolumn{7}{l}{\text{otherwise}}
        \end{array}
    \end{aligned}
    \label{fig:KP}
    \end{equation}
    where by deriving the definition of $\KP_i(w, \sigma)$
    directly from the definition of $f_\sigma^{\kw{dft}}(cd)$,
    there is a correspondence between the definitions,
    which is what we call ``coupling''.

    Formally,
    we claim  that the FSA-based supervisors 
    constructed as $(G^{obs}_i, \KP_i)$,
    satisfy
    \cref{eq:solve-2,eq:solve-3}.
    Since for any string $s \in L(E)$,
    the two propositions $s\sigma \in L(E)$ and $s\sigma \not\in L(E)$ are mutually exclusive,
    we only need to show one of \cref{eq:solve-2,eq:solve-3},
    the other vacuously holds true.

    We perform a case analysis over inference\-/observability \cref{eq:inf-obs}
    and show that in each case one of \cref{eq:solve-2,eq:solve-3} holds.
    For each case,
    consider only $w$ where $w_e \in Q^E$ and $\sigma$ such that $\sigma \in \Sigma_c$.
    We will also assume that $\sigma$ can happen at state $w$,
    that is, assume $\sigma_G$:
    if it is not physically possible for $\sigma$ to happen at state $w$,
    we are not obligated to make any decision.

    Suppose \cref{eq:inf-obs-1} holds for some $i$, but \cref{eq:inf-obs-2} does not hold for any $i$.
    Then we have $e$ and supervisor $i$ issues $\cdon$.
    Also none of the supervisors can issue $\cdoff$,
    since otherwise contradicting $e$.
    Regardless which default action we choose for $\sigma$ (similarly hereinafter unless explicitly stated otherwise),
    the joint decision is $\fde$,
    i.e., $f_\nset(s, \sigma) = \fde$,
    thus \cref{eq:solve-2} holds.

    Suppose \cref{eq:inf-obs-2} holds for some $i$,
    but \cref{eq:inf-obs-1} does not hold for any $i$.
    The result follows in an analogous way to the previous case.

    Suppose both \cref{eq:inf-obs-1} and \cref{eq:inf-obs-2} hold true for not necessarily the same $i$.
    In this case we have $e \land d$,
    which implies $\lnot \sigma_G$.
    Thus \cref{eq:solve-2,eq:solve-3} both hold vacuously.
    Informally,
    since $\sigma$ cannot happen at state $w$,
    we have no obligation to make a decision regarding $\sigma$.
    This is the reason that the control protocol is not defined to be total: it lacks the case
    $(I, w) \models K_i e \land K_i d$,
    and the fusion rule does not consider the case where both $\cdon$ and $\cdoff$ are issued,
    unlike that of \citet{Ricker2007}.
    This also suggests that our expression of the fusion rule automatically ensures \emph{control-nonconflicting} \citep{Yoo2004} supervisors.
    
    Then consider the following cases,
    where \cref{eq:inf-obs-1} and \cref{eq:inf-obs-2} both do not hold.
    
    Suppose \cref{eq:inf-obs-3} holds for some $i$,
    but \cref{eq:inf-obs-4} does not hold for the same $i$.
    Then we have $\underline{e} \implies O e$ 
    and supervisor $i$ issues decision $\cdwoff$.
    Again, consider only cases where $\sigma_G$ holds.
    If $\sigma_E [= \underline e]$,
    then for some $j \in \nset_\sigma$ other than $i$,
    we have $K_j e$
    and thus supervisor $j$ issues $\cdon$.
    Also no other supervisor issues $\cdoff$,
    since otherwise this would contradict $e$.
    Hence by the fusion rule,
    the joint decision is $\fde$,
    thus \cref{eq:solve-2} holds.
    If, however, $\lnot \sigma_E [= d]$,
    then no supervisor issues $\cdon$,
    otherwise this would contradict $d$.
    Hence either some other supervisor issues $\cdoff$,
    in which case the fused decision would be $\fdd$;
    or there is no such supervisor issuing $\cdoff$,
    but some supervisor $j$ other than $i$ issuing $\cdwon$,
    in which case,
    it must be $\underline{d} \implies \lOr_{\stackrel{j \in \nset}{j \ne i}} O d)$.
    Since we have $d$,
    there must be a third supervisor issuing $\cdoff$,
    which overrides both conditional decisions
    (since there is no $\cdon$ issued),
    and hence the fused decision would be $\fdd$;
    lastly, if no $\cdoff$ nor $\cdwon$ is issued,
    the fused decision would also be $\fdd$.
    Thus $\fdd$ is issued in all cases,
    and \cref{eq:solve-3} holds.
    
    Suppose \cref{eq:inf-obs-4} holds for some $i$,
    but \cref{eq:inf-obs-3} does not hold for the same $i$.
    The result follows in an analogous way to the previous case.

    Suppose both \cref{eq:inf-obs-3} and \cref{eq:inf-obs-4} hold true for the same $i$.
    In this case supervisor $i$ would issue decision $\cdabs$.
    Also we have
    $
    [\sigma_E \implies O e]
    \land
    [\lnot\sigma_E \implies O d]$
    (since $\sigma_G$ has been assumed).
    In the case of $\sigma_E$,
    for some $j_1 \in \nset_\sigma$ other than $i$,
    we have $K_{j_1} e$,
    i.e., \cref{eq:inf-obs-1} holds for $j_1$,
    and hence $j_1$ issues $\cdon$.
    By the fusion rule,
    the joint decision is $\fde$,
    thus \cref{eq:solve-2} holds.
    In the case of $\lnot\sigma_E$,
    for some $j_2 \in \nset_\sigma$ other than $i$,
    we have $K_{j_2} d$,
    i.e., \cref{eq:inf-obs-2} holds for $j_2$,
    and hence $j_2$ issues $\cdoff$.
    By the fusion rule,
    the joint decision is $\fdd$,
    thus \cref{eq:solve-3} holds.
    Notice the choice of instantiations of $j_1$ and $j_2$ are arbitrary;
    in particular there is no requirement that $j_1 \ne j_2$.
    Previously in the work of \citet{Yoo2004},
    supervisor $i$ would issue the \textbf{don't know} decision in this case,
    which is semantically identical to $\cdabs$ as specified by the fusion rule.
    However our argument indicates that in this case 
    supervisor $i$ knows that whatever has to be done about $\sigma$ has already been taken care of by supervisors $j_1, j_2$
    hence it actively decides not to engage in shaping the fused decision.
    Given this interpretation,
    it is, therefore, more suitable to change the \textbf{don't know} decision in this case to $\cdabs$.

    Suppose \cref{eq:inf-obs-3} but not \cref{eq:inf-obs-4} holds true for some $i_1$,
    while not \cref{eq:inf-obs-3} but \cref{eq:inf-obs-4} holds true for some other $i_2$.
    I.e., supervisor $i_1$ issues $\cdwoff$ while supervisor $i_2$ issues $\cdwon$.
    With similar reasoning as the previous case,
    it is always the case that some third supervisor would issue $\cdon$ or $\cdoff$
    (since $\sigma_G$ has been assumed),
    either way, we derive one of \cref{eq:solve-2,eq:solve-3}.
    This highlights why the fusion rule need not consider the situation
    when both $\cdwoff$ and $\cdwon$ are issued,
    and yet neither $\cdon$ nor $\cdoff$ is issued.

    Finally, if for some $i$ \cref{eq:inf-obs-1,eq:inf-obs-2,eq:inf-obs-3,eq:inf-obs-4} do not hold but only \cref{eq:inf-obs-5} holds,
    then we have $\phi_\sigma$,
    which is either $\lnot \sigma_G \lor \sigma_E$ or $\lnot \sigma_E$.
    
    If $\phi_\sigma = \lnot \sigma_G \lor \sigma_E$,
    since we do not consider the case where $\lnot \sigma_G$,
    it must be $\sigma_E$.
    Hence it is impossible for some supervisors to issue $\cdoff$.
    If some supervisor $j_1$ other than $i$ issues $\cdwoff$,
    then since $\sigma_E$, there must be a third supervisor $j_2$ issues $\cdon$.
    If some supervisors other than $i$ issue $\cdwon$,
    then $\cdoff$ and $\cdwoff$ cannot be issued.
    In the two cases above,
    to get a fused decision $\cdon$,
    by the fusion rule,
    we need $i$ to issue $\cdabs$,
    and \cref{fig:KP} is deliberately constructed to ensure this.
    However,
    if all supervisors issue $\cdabs$,
    then in order to have $\fde$,
    we choose $\kw{dft} = \fde$.
    In any case, we have shown that \cref{eq:solve-2} holds.

    In the case of $\phi_\sigma = \lnot \sigma_E$,
    by a similar argument to the aforementioned argument,
    we can let $i$ issue $\cdabs$,
    and choose $\kw{dft} = \fdd$,
    and thus \cref{eq:solve-3} holds.

    We have now exhausted all cases of \cref{eq:inf-obs},
    and derived \cref{eq:solve-2,eq:solve-3} as desired.

    Notice that inference-observability ensures consistent choice of $f_\sigma$.

    A final comment on the case where
    for some $i$ \cref{eq:inf-obs-1,eq:inf-obs-2,eq:inf-obs-3,eq:inf-obs-4} do not hold but only \cref{eq:inf-obs-5} holds:
    this is the case where it might have been more accurate to denote the control decision
    as \textbf{don't know} instead of $\cdabs$.
    However, as discussed,
    since the fusion rule does not have to treat differently
    the decisions \textbf{don't know} and $\cdabs$,
    we have elected to not include both \textbf{don't know} and $\cdabs$,
    but use only $\cdabs$.

    We have opted for a less rigorous description of the control policies
    $\KP_i$ for conciseness,
    hence we elaborate further.
    We defined the control policies $\KP_i$
    over the Kripke Structure $W$,
    which is essentially the automaton $G' = G \times P_1(G) \times \dots \times P_n(G)$.
    However,
    the automaton $G'$ contains unobservable events
    and hence is not suitable to be used as supervisors.
    The solution is simple:
    use the automata $P_i(G)$ for the supervisors and
    construct the control policies
    % We constructed the FSA-based supervisors as $(G^{obs}_i, \KP_i)$,
    % where the FSA $G^{obs}_i$ have state spaces $W_i$ but $\KP_i$
    % take arguments from $W$ instead of $W_i$.
    % This gap can be filled by constructing the control policies
    $\KP'_i : W_i \times \Sigma_{c, i} \to \CD$
    so that $\KP'_i(w_i, \sigma) = \KP_i(w, \sigma)$
    whenever the $i$'th component of $w$ is $w_i$.
    This construction is well-defined
    since for all $w$ whose $i$'th component is $w_i$,
    $\KP_i(w, \sigma)$ takes the same value.
\end{pf}

The choice of default decision in the proof above indicates how one
can obtain a partition $\Sigma_c = \Sigma_{c,e} \union \Sigma_{c,d}$.
Readers familiar with the work of \citet{Yoo2004} might have noticed that 
there could be multiple such partitions so that the language is inference-observable
(or conditionally co-observable \citep{Yoo2004}, since one can directly prove that these two conditions are equivalent).
With the epistemic expression, we are able to give the following interpretation of this phenomenon:
these partitions differ exactly on the events for which there is no world where all supervisors abstain from voting.

The proof of \cref{thm:inf-obs-extended} shows how a proof of \cref{thm:inf-obs-corrected} can be constructed:
with the expression of inference\-/observability being directly coupled with the control policy,
one simply performs a case analysis over how some supervisor $i \in \nset_\sigma$ makes its decision and reuse most of the paragraphs in the proof of \cref{thm:inf-obs-extended}.
% Similar result can be added to \cref{sec:multi}.

One can see how to modify the expression of inference\-/observability (\cref{eq:inf-obs})
when some of the control decisions we discussed are no longer permitted.
It is particularly interesting to see that with only $\{\cdoff, \cdabs\}$
we arrive at C\&P co\-/observability,
and with only $\{\cdon,\cdabs\}$,
at D\&A co\-/observability \cite{Yoo2004}.
In these two cases,
if the Kripke structures are constructed as $\overline{I}$ instead of $I$,
we have instead strong C\&P or strong D\&A co-observabilities \citep{Takai2005}.
See \citet{Ean2021a} for details.

% \end{verArxiv}

% Up until this point,
% we have only discussed inference\-/observability for specific implementations of $L(G)$ and $L(E)$.
% It is then fair to ask whether inference\-/observability is implementation independent,
% i.e., whether it is appropriate to discuss inference\-/observability of a prefix-closed regular language $L$.

% Formally, the \emph{implementation independence} of inference\-/observability is expressed as follows.
% Given two pairs of implementations of the plant/legal behaviours $(G_1, E_1)$ and $(G_2, E_2)$ such that $L(G_1) = L(G_2)$, $L(E_1) = L(E_2) = L$,
% and $E_i$ is arranged as a subautomaton of $G_i$ for both $i = 1, 2$,
% then is the inference\-/observability of $E_1$ w.r.t $G_1$ equivalent to that of $E_2$ w.r.t. $G_2$.

% The answer is positive.
% This follows from the fact that conditional co\-/observability \cite{Yoo2004} is equivalent to inference\-/observability.
% Since conditional co\-/observability can be verified against an arbitrary implementation,
% it is implementation independent.
% Hence inference\-/observability is also implementation independent.

\section{Using the Epistemic Logic Formalism to Revise Problem Requirements}

We provide an example of a non-inference-observable language.
We then point out how epistemic expression of inference\-/observability reveals what prevents the language from being inference-observable.
Finally we discuss how inference-observable languages can be obtained based on the non-inference-observable one,
including a sublanguage.

Consider the following example:
given a plant $G$ with event set $\Sigma = \set{\alpha_1, \alpha_2, \beta_1, \beta_2, \gamma, \mu}$,
do there exist two supervisors,
with observed event sets $\Sigma_{1, o} = \set{\mu}$, $\Sigma_{2, o} = \set{\beta_1, \beta_2}$,
and controlled event sets $\Sigma_{1, c} = \Sigma_{2, c} = \set{\gamma}$,
such that the given language $L(E)$ is inference-observable?

\cref{fig:non-inf-obs} depicts the automaton $G' = G \times P_1(G) \times P_2(G)$.
Since $G'$ and $G$ happens to be isomorphic in this example,
we do not draw $G$ separately.
The language $L(E)$ is marked by states with double borders.

% https://personal.sron.nl/~pault/
% Vibrant qualitative colour scheme
\newcommand{\TL}{\textcolor[HTML]{0077bb}{\ensuremath{\pmb{\rm{M}_0}}}}
\newcommand{\TR}{\textcolor[HTML]{33bbee}{\ensuremath{\pmb{\rm{M}_1}}}}
\newcommand{\WL}{\textcolor[HTML]{ee7733}{\ensuremath{\pmb{\rm{B}_0}}}}
\newcommand{\WU}{\textcolor[HTML]{cc3311}{\ensuremath{\pmb{\rm{B}_1}}}}
\newcommand{\WD}{\textcolor[HTML]{ee3377}{\ensuremath{\pmb{\rm{B}_2}}}}

\begin{figure*}[htbp]
  \centering
  \includegraphics[width=1\linewidth]{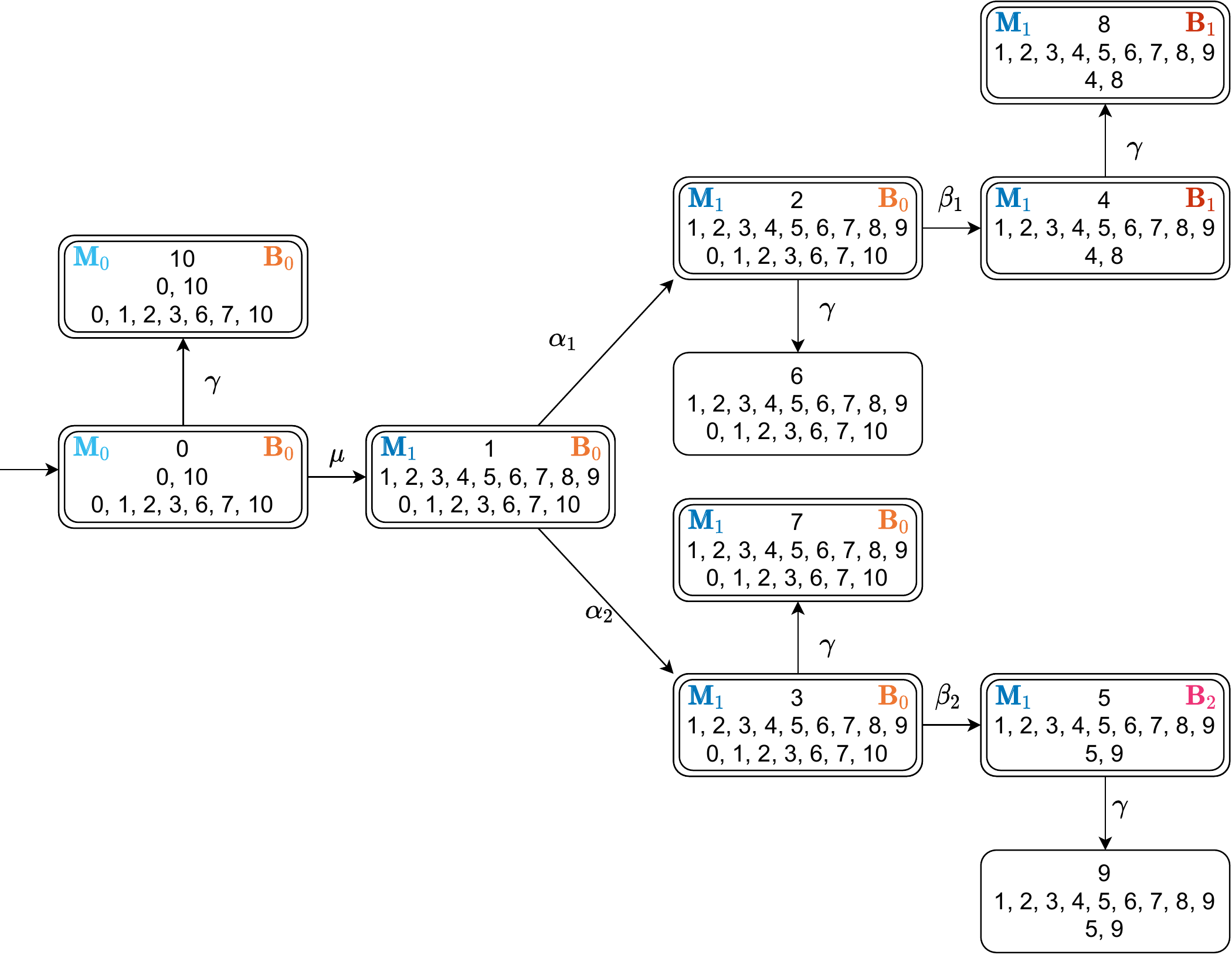}
  \caption{
    The automaton $G' = G \times P_1(G) \times P_2(G)$.
    A state $(q_G, q^{obs}_1, q^{obs}_2)$ is
    represented in the figure with $q$, $q^{obs}_1$, $q^{obs}_2$ stacked vertically in that order.
    The equivalence classes are marked according to the following rule:
    a state is marked at the upper left (resp. upper right) corner
    according to its containing equivalence class formed by the accessibility relation $\sim_1$ (resp. $\sim_2$);
    the symbols for the equivalence classes are deliberately chosen,
    so, for instance, the states supervisor 2 thinks the plant could be in after it sees $\beta_1$ are in the equivalence class $\WU$.
  }
  \label{fig:non-inf-obs}
\end{figure*}

Let us focus on $\gamma$ since it is the only controllable event.
Hence we focus on states $0$, $1$, $2$, $3$, $4$, $5$,
since these are the states where $\gamma$ can happen.

In state $4$ (resp. $5$),
supervisor 2 can enable (resp. disable) $\gamma$.
In state $0$, supervisor 1 can enable $\gamma$.
But in states $2$, $3$, which are indistinguishable to both supervisors,
since they are both in the same equivalence classes
($\TR$ for supervisor 1 and $\WL$ for supervisor 2),
neither supervisor 1 nor 2 can control $\gamma$ unambiguously.
Hence the language $L(E)$ is not inference-observable.

The representation of $G'$ and the epistemic interpretation of conditional control decisions provides guidance for how to modify the control requirement to obtain an inference-observable language.

If we are looking for a sublanguage,
we can only make legal states illegal but not vice versa.
By our previous analysis,
at least one supervisor is able to make a correct control decision unambiguously in states $S = \{0, 1, 4, 5, 7, 8, 10\}$,
hence all we need to worry about are the states in the set $\TR - S = \WL - S = \{2, 3\}$.
To resolve the conflict that $\gamma$ is legal at state $3$ but illegal at state $2$,
we can make state $7$ illegal.

To see how making state $7$ illegal gives an inference-observable sublanguage,
let's look at states in $\TR$ and $\WL$.
At states in $\TR$,
$\gamma$ is illegal at states $2, 3, 5$
but is legal at state $4$.
With only binary control decisions,
supervisor 1 cannot possibly make an unambiguous decision.
We can see that supervisor 2 is in a similar situation by examining states in $\WL$.

However, with the ability to infer the knowledge of other supervisors
and the conditional decisions at their disposal,
the desired control requirement can be achieved.
Suppose that supervisor 1 is an intelligent being,
and let's imagine how the intelligent being may attempt to solve the dilemma.
If supervisor 1 were to try ``guessing'' the legality of $\gamma$ after it sees $\mu$,
it would realize,
that even the guess ``$\gamma$ is illegal'' is not always correct,
i.e., it is false at exactly state $4$,
by knowing that the other supervisor can unambiguously enable $\gamma$ if the plant is indeed at state $4$,
supervisor 1 is then able to focus on only the rest of the states in $\TR$,
and fortunately,
its guess is correct in all of them,
hence supervisor 1 can confidently disable $\gamma$ at states in $\TR$ unambiguously,
knowing its mistake would be corrected by the other supervisor.
Similar reasoning is also carried out by supervisor 2.

The design of the fusion rule is exactly to allow the correction of mistakes.
A $\cdwoff$ is issued by a supervisor knowing that if disabling the event is incorrect then another supervisor can correct the first supervisor by a definite $\cdon$ decision.

Formally, with state $7$ made illegal,
states $\set{2, 3}$ are unambiguous.
However, since the set $\set{2, 3}$ is a proper subset of both $\TR$ and $\WL$,
and states in both sets $\TR$ and $\WL$ remain ambiguous,
the conditional decision, i.e., $\cdwoff$ has to be issued at states in the set $\TR$ (resp. $\WL$) by supervisor 1 (resp. supervisor 2).

If we are open to not necessarily a sublanguage, we can also make state $6$ legal too.
By similar reasoning as we just did,
supervisor 1 should issue decision $\cdwon$ at states $2$, $3$;
and supervisor 2 can issue decision $\cdon$ at states in the set $\WU$,
since this set is no longer ambiguous.

\section{Conclusion}

In this paper,
we discuss how decentralized control with non-binary control decisions \citep{Yoo2004} can benefit from the use of epistemic logic.

We point out
that epistemic logic can be used to discuss not only some specific classes of DSCOP \citep{Ricker2000,Ricker2007},
but also more universally.
We demonstrated this by showing how epistemic logic
formally encapsulates the expression $\All{s, s_i} P_i(s) = P_i(s_i) \implies \dots$ (or similarly $P_i^{-1}P_i(\cdot)$) used ubiquitously in discussions of DSCOP,
and by informally personifying supervisors so that we can understand and discuss the control problem with an anthropomorphic perspective and language.

We deliberately coupled
the epistemic expression characterizing the class of DSCOP discussed by \citet{Yoo2004}
and the expression describing the control policies.
This line-by-line coupling allows us to use the same expression throughout the discussions of proving necessary and sufficient conditions, of describing the algorithm to construct the supervisors, and of verifying the correctness of the algorithm.

From the forgoing discussions,
we would expect other decentralized control or diagnosis conditions could be treated in a comparable fashion.
For instance,
consider the work of \citet{Kumar2007},
which is more general than that of \citet{Yoo2005}.
We develop our epistemic expressions based on \citet{Yoo2005}
because it is simpler and thus we are able
to demonstrate our key ideas without more complex
(yet not conceptually different) technical development.
The same principles demonstrated here could apply to \citet{Kumar2007} as well.
The only technical difference is that one would need to use a finer,
(possibly infinite) string-based Kripke structure as described by \citet{Ricker2000},
along with a corresponding definition of relations $\sim_i$.

Casting the decentralized problem the way we did makes it easier to
understand the reasoning behind various control decisions.
We believe that one advantage of our framework is that
in trying to come up with solutions to future DES problems,
this framework can aid in going directly from a working supervisor solution
to the necessary and sufficient conditions that would match such a solution.
Moreover,
if the constraints of some given problem are not met
(and hence that problem is not solvable as is using decentralized control),
our model makes it more apparent how to alter the constraints
in a way that is meaningful for the application at hand.

% We have to emphasize that inference\-/observability is discussed for a particular pair of implementations $G$ and $E$ of the plant/legal behaviours.
% In other words, we do not discuss whether a certain prefix-closed regular language $L$ is inference-observable,
% independent of its implementation.

% The epistemic interpretation provides a guidance to obtain a inference-observable sublanguage.
% The discussion indicates,
% by judiciously make legal states illegal,
% we can obtain a maximal inference-observable sublanguage.
% However the sublanguage is not necessarily supremal.

\section*{Acknowledgements}
The research described in this paper was undertaken at Queen's University,
which is situated on traditional Anishinaabe and Haudenosaunee territory.
The research was inspired by and supported through an NSERC CRD\-/DND project
with General Dynamics Land Systems\--Canada and Defence Research and Development Canada.

\section*{Conflicts of Interest Statement}
The authors declare that they have no conflict of interest.

\bibliographystyle{my-natbib-IEEEtranN}
\bibliography{IEEEtranBSTCTL, ../bib/DES.bib, local.bib}

\vfill

\end{document}